\documentclass[lettersize,journal]{IEEEtran}
\usepackage{amsmath,amsfonts}
\usepackage{algorithmic}
\usepackage{algorithm}
\usepackage{array}
\usepackage[caption=false,font=normalsize,labelfont=sf,textfont=sf]{subfig}
\usepackage{textcomp}
\usepackage{stfloats}
\usepackage{url}
\usepackage{verbatim}
\usepackage{graphicx}
\usepackage{cite}
\usepackage{amssymb}
\usepackage{xcolor}
\usepackage{booktabs}
\usepackage{makecell}
\usepackage{ulem}
\usepackage{tabularx}
\usepackage[table]{xcolor}
\usepackage{multirow}

\newcolumntype{L}[1]{>{\raggedright\arraybackslash}p{#1}}
\newcolumntype{Y}{>{\centering\arraybackslash}X}
\begin{document}

\title{Earth-Agent-Pro: Towards Real-World Full-Chain Earth Observation with Agents}

\author{
    Zhutao Lv\textsuperscript{\rm 1,2},
    Chenhao Dang\textsuperscript{\rm 3,4}\textsuperscript{*}, 
    Yi Feng\textsuperscript{\rm 5}\textsuperscript{*}, 
    Yanpei Gong\textsuperscript{\rm 6}\textsuperscript{*}, 
    Xiaolei Wang\textsuperscript{\rm 2},    \\
    Junyan Ye\textsuperscript{\rm 2},
    Conghui He\textsuperscript{\rm 4},
    Weijia Li\textsuperscript{\rm 1,4}\textsuperscript{\dag} 

    \textsuperscript{\rm 1}Tsinghua Shenzhen International Graduate School, Tsinghua University,
    \textsuperscript{\rm 2}Sun Yat-Sen University, \\
    \textsuperscript{\rm 3}Shanghai Jiao Tong University,
    \textsuperscript{\rm 4}Shanghai Artificial Intelligence Laboratory, \\
    \textsuperscript{\rm 5}Tianjin University,
    \textsuperscript{\rm 6}Harbin Institute of Technology
    \thanks{
        \textsuperscript{*}Equal contribution. \quad 
        $^\dagger$Corresponding author.
    }
}



\maketitle

\begin{abstract}
Real-world Earth observation (EO) analysis requires agents to be capable of translating high-level scientific questions into executable workflows for acquiring observations, preparing data, performing domain computations, and deriving conclusions from runtime evidence. However, existing EO agents and benchmarks generally cover only part of this chain: agents typically start from supplied observations, whereas benchmarks typically provide prepared inputs or candidate answers, leaving full-chain open-world EO execution largely untested. We present Earth-Agent-Pro, an execution-adaptive Plan-and-Execute framework that uses expert-authored skills to constrain workflow planning and runtime tool use. It records planned steps, accepted evidence, and their dependency links in workflow-centered structured memory, then repairs only the affected workflow suffix when runtime evidence invalidates a step. Role-specific training adapts the underlying large language model (LLM) through separate planner and executor adapters: sequence-level supervised fine-tuning targets planner workflow composition, while node-level group relative policy optimization with locally verifiable rewards targets executor tool-argument grounding. To evaluate full-chain open-world EO execution, we introduce Earth-Bench-Pro, a 744-question EO agent benchmark that instantiates 248 expert-curated task cores under three matched regimes. Across RGB imagery, spectral observations, and remote sensing products, its 248 Open-World Execution questions pair high-level requests with runtime data requirements, executable trajectories, and open-ended answers grounded in execution evidence. Under a shared GPT-5 backbone, Earth-Agent-Pro reaches 66.13\% LLM-as-Judge accuracy, outperforming ReAct by 20.95 points in LLM-as-Judge accuracy and 24.44 points in Tools-In-Order. Jointly tuning the Planner and Executor adapters raises Qwen3.5-9B LLM-as-Judge accuracy from 38.31\% to 50.00\%, an 11.69-point gain over the untuned configuration. Planning-only evaluation and execution with the reference workflow further show that the adapters improve workflow composition and argument grounding, respectively. In conclusion, Earth-Bench-Pro and Earth-Agent-Pro support evaluation and training of EO agents from user requests to evidence-grounded scientific answers. The code
and datasets of this work will be released soon.

\end{abstract}


\section{Introduction}
\label{sec:introduction}

\begin{figure*}[!t]
	\centering
	\includegraphics[width=0.95\textwidth]{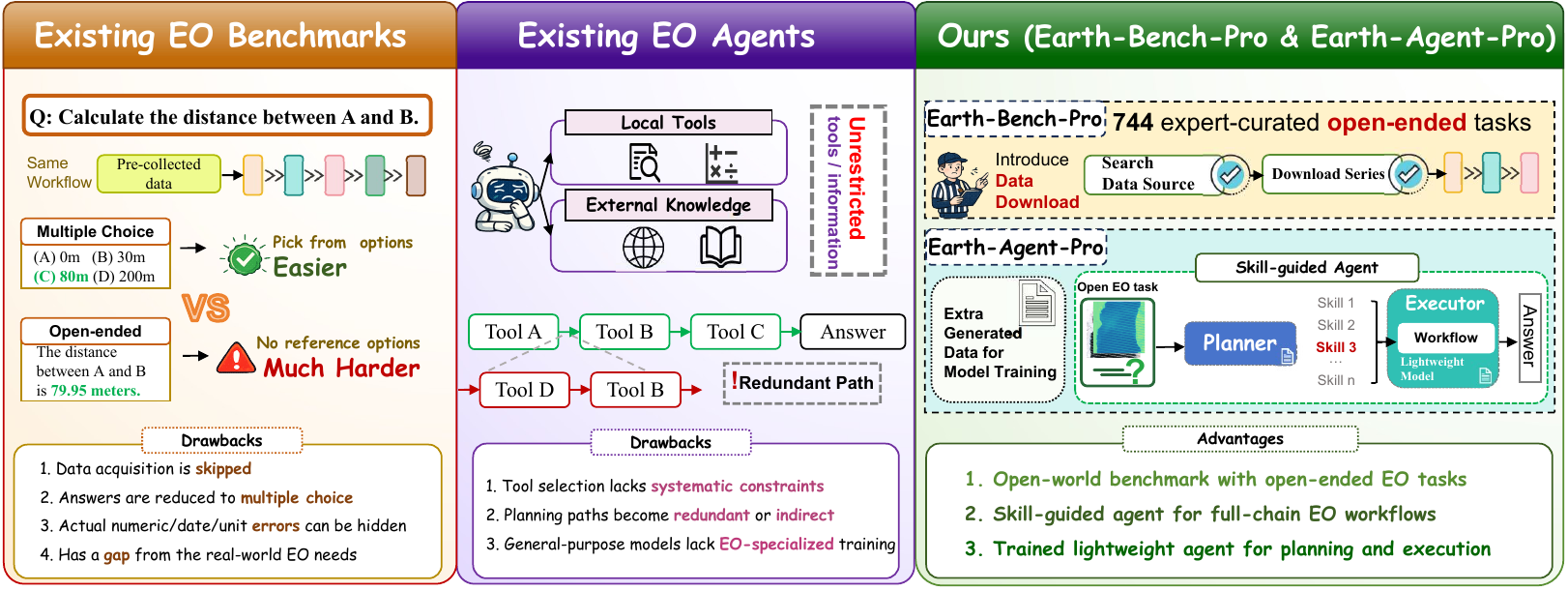}\\
	\caption{\textbf{Overview of our work.}  Existing EO benchmarks rely on pre-collected data and multiple-choice evaluation, while existing EO agents suffer from unconstrained tool use and redundant planning. Our work introduces an open-ended EO benchmark (Earth-Bench-Pro) and a trained skill-guided agent (Earth-Agent-Pro) for full-chain EO task planning and execution.}\label{fig1}
\end{figure*}

Earth observation (EO) turns satellite and aerial measurements into quantitative evidence for agricultural monitoring, disaster assessment, urban analysis, and environmental change research~\cite{transon2018survey, ye2025satellite, anderson2017earth, zhou2025urbench, kokkoris2024role, li2024crossviewdiff, li2024roadcorrector, brown2025alphaearth, ye2024skydiffusion, liu2020large, ye2024sg, li20243d, li2023joint, li2020integrating,li2023omnicity}. Producing such evidence from an EO question remains expert-intensive~\cite{Sudmanns02012018, WEISE2020111892, FENG2025104825}. An EO analysis begins by matching the target phenomenon, region, and period to suitable observations. Analysts then choose the data source, product, bands, and sampling parameters before acquiring and preprocessing the data. The prepared observations enter a domain-specific computation pipeline that yields the requested scientific quantities. Completing this workflow requires expertise in sensors, data products, geospatial processing, and domain science, which raises the entry barrier for users outside the EO community and limits access to reproducible analysis.

Recent work has expanded EO capabilities beyond interpreting supplied observations to include multistep analysis execution. Vision-language models interpret supplied images and image sequences~\cite{kuckreja2024geochat,muhtar2024lhrs,shu2026terrascope}, while tool-augmented agents select operations, execute code, and construct multistep workflows~\cite{shabbir2025thinkgeo,kao2026towards,li2025designing,shabbir2026openearthagent, dang2026skiller}. Benchmarks now span fixed-input question answering, multistep geographic information system (GIS) tool use, and open-environment pipeline construction~\cite{lobry2020rsvqa,li2024vrsbench,krechetova2025geobenchx,zhao2026openearth}. Together, these methods and benchmarks cover more of the request-to-answer chain, but evaluations differ in their scientific tasks, input assumptions, and answer formats. Measuring progress toward autonomy therefore requires a matched evaluation that removes expert scaffolding while keeping the scientific objective fixed.

Our prior Earth-Agent and Earth-Bench provide the task and evaluation foundation for this comparison~\cite{feng2025earth}. Earth-Agent connects a language model to EO tools through the Model Context Protocol, while Earth-Bench evaluates final answers and reference trajectories across red-green-blue (RGB) imagery, spectral observations, and remote sensing products. This pairing evaluates both the scientific result and the execution path. Earth-Bench, however, begins with expert-prepared local inputs and multiple-choice questions, leaving runtime data discovery and open-ended answer generation outside its evaluation boundary. We build on this foundation by preserving each of Earth-Bench's 248 scientific objectives as a fixed task core and moving runtime data discovery, acquisition, and open-ended answer generation into the evaluated workflow.

The resulting request-to-answer setting, which we call \textit{open-world EO execution}, starts from a request that specifies a phenomenon, region, and period. The agent must locate runtime inputs, acquire suitable observations when required, prepare them, perform domain computations, and derive an answer from runtime evidence. Files, intermediate artifacts, and tool results appear only during execution, so later operations depend on accepted upstream evidence. The agent must track workflow dependencies, ground each operation in the evolving state, and revise affected future operations when evidence invalidates the current plan.

\begin{figure}[t]
    \centering
    \includegraphics[width=0.95\linewidth]{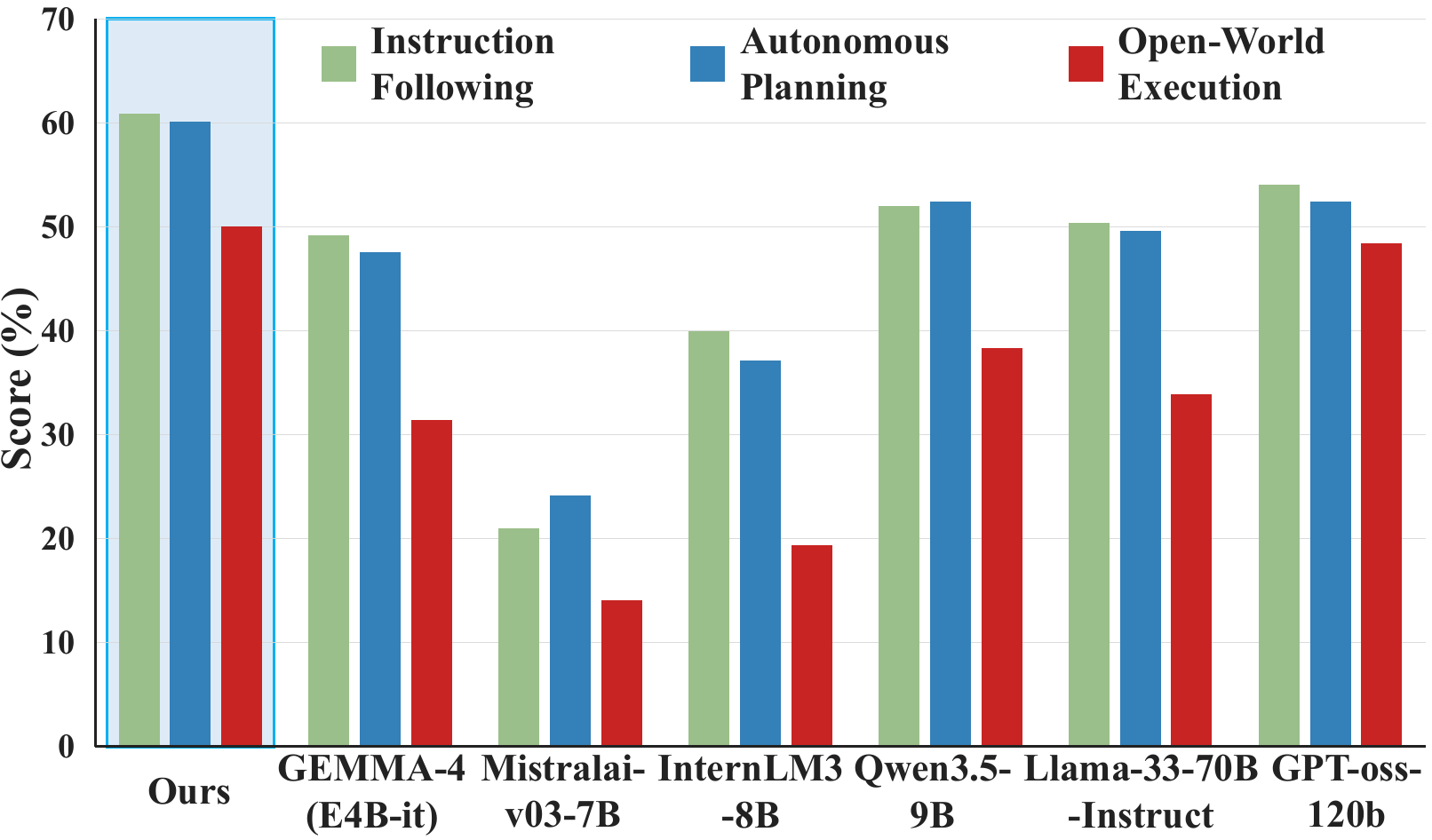}
    \caption{
    LLM-as-Judge accuracy across Earth-Bench-Pro's three regimes.
    \textbf{Ours} is our role-specialized Qwen3.5-9B.
    }
    \label{fig:benchmark_difficulty}
\end{figure}

We develop \textbf{Earth-Bench-Pro} and \textbf{Earth-Agent-Pro} as a paired benchmark and agent for open-world EO execution. Earth-Bench-Pro holds each scientific objective fixed across matched regimes that progressively remove the reference workflow and prepared inputs. Earth-Agent-Pro separates workflow planning from runtime operation grounding and constrains both stages with expert-authored skills. Workflow-centered structured memory links pending operations to accepted evidence, allowing the agent to preserve a valid prefix and repair only an invalidated workflow suffix. The training pipeline specializes the Planner for workflow composition and the Executor for runtime argument grounding.

Earth-Bench-Pro exposes a consistent difficulty gap: Open-World Execution remains the hardest regime across backbones (Figure~\ref{fig:benchmark_difficulty}). On Earth-Bench-OW, with a shared GPT-5 backbone, tool set, and skill set, Earth-Agent-Pro reaches 66.13\% accuracy under a large language model (LLM) judge, 20.95 percentage points above ReAct, and leads ReAct by 24.44 points on Tools-In-Order. Combining the Planner and Executor adapters raises Qwen3.5-9B LLM-as-Judge accuracy by 11.69 points; planning-only and oracle-plan evaluations separately measure workflow composition and argument grounding.

Our contributions are:
\begin{itemize}
    \item \textbf{Open-world execution task and benchmark.} We formulate open-world EO execution and operationalize it with 248 Open-World Execution questions in Earth-Bench-Pro. Product and Spectrum tasks require runtime observation acquisition and preparation, and all questions require open-ended answers grounded in execution evidence. Together with 496 matched Instruction-Following and Autonomous-Planning questions, the released suite contains over 700 evaluation questions.

    \item \textbf{Open-world EO agent framework.} Earth-Agent-Pro uses expert skill guidance to constrain workflow construction and runtime tool use, while its execution-adaptive dynamic workflow maintains the plan, accepted evidence, and their dependencies. Under a shared backbone, tool set, and skill set, Earth-Agent-Pro outperforms ReAct, AFlow, and OpenEarthAgent on final-answer and workflow-order metrics in controlled comparisons.

    \item \textbf{Dependency-preserving data and role training.} From 248 seed workflows, we construct 2,480 dependency-preserving trajectories and combine them with examples from the public OpenEarthAgent training set~\cite{shabbir2026openearthagent} to form a balanced mixture for Planner SFT and Executor GRPO. Combining the two adapters raises Qwen3.5-9B LLM-as-Judge accuracy by 11.69 points on Earth-Bench-OW.
\end{itemize}

\section{Related Work}

\begin{figure*}[!t]
	\centering
	\includegraphics[width=0.95\textwidth]{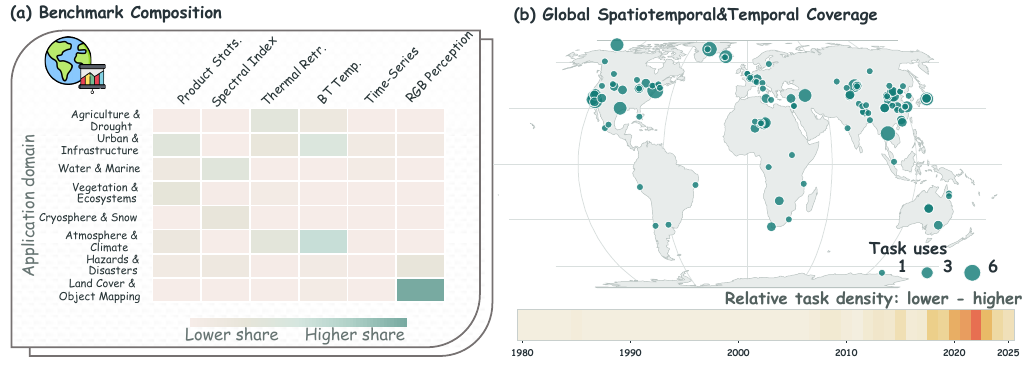}\\
	\caption{\textbf{Earth-Bench-Pro Overview.}
(a) Task composition across application domains and task families.
(b) Global spatial distribution and temporal coverage.}\label{benchmark_overview}
\end{figure*}

\subsection{Earth Observation Benchmarks and Datasets}

Earth observation (EO) datasets provide perception and multimodal evaluation on fixed inputs~\cite{wang2025omniearth}. RSVQA \cite{lobry2020rsvqa} and VRSBench \cite{li2024vrsbench} assess visual question answering; GeoBench-VLM \cite{danish2025geobench} and CHOICE \cite{an2026choice} evaluate geospatial multimodal capabilities; VLEO-Bench \cite{zhang2024good}, XLRS-Bench \cite{wang2025xlrs}, URBench \cite{zhou2025urbench}, DisasterM3 \cite{wang2026disasterm3}, and TerraScope \cite{shu2026terrascope} extend evaluation to fine-grained, ultra-high-resolution, urban, disaster, and pixel-grounded reasoning. These benchmarks mainly score final outputs rather than agent interactions, tool-use trajectories, or intermediate artifacts. Complementary EO benchmarks target specific agentic operations: RescueADI \cite{liu2025rescueadi} evaluates sequential disaster interpretation; GeoLLM-Engine \cite{singh2024geollm} provides analyst-style API and interface interactions; GeoBenchX \cite{krechetova2025geobenchx} tests multistep GIS tool calling; AEOS-Bench \cite{wang2026towards} considers constrained constellation scheduling. Agent-oriented benchmarks formalize these capabilities in their protocols: ThinkGeo \cite{shabbir2025thinkgeo} assesses tool use, Earth-Bench \cite{feng2025earth} evaluates trajectory-level EO tasks, and HTAM/GeoPlan-Bench \cite{li2025designing} targets hierarchical task abstraction. More recently, EO-Gym \cite{ma2026eo} and TerraBench \cite{nguyen2026terrabench} introduce interactive heterogeneous data; GeoMMBench \cite{xiao2026geommbench} and TerraLogic \cite{yan2026terralogic} target complex geoscientific reasoning; and GeoAgentBench \cite{yu2026geoagentbench} and DORA \cite{wang2026dora} assess GIS execution verification and emergency geospatial reasoning. OpenEarth-Bench \cite{zhao2026openearth} evaluates open-environment EO pipelines and tool creation, whereas NASA-EO-Bench \cite{yu2026nasaeobench} focuses on data and tool discovery. Yet evaluations often start from curated cases or treat discovery separately, leaving open the joint assessment of data-source selection, acquisition, intermediate evidence, and final natural-language answers.

\subsection{Earth Observation Methods}

\textbf{MLLM-based EO Methods.} EO representation and perception methods, including SpectralGPT \cite{hong2024spectralgpt}, RingMo-Aerial \cite{diao2025ringmo}, REST \cite{chen2025rest}, RingMoE \cite{bi2025ringmoe}, CrossEarth \cite{gong2025crossearth}, and HyperSIGMA \cite{wang2025hypersigma}, have strengthened representation learning and visual interpretation across spectral, aerial, and hyperspectral observations~\cite{zhang2026a2mae}. In parallel, RSGPT \cite{hu2025rsgpt}, SkySenseGPT \cite{luo2024skysensegpt}, GeoChat \cite{kuckreja2024geochat}, LHRS-Bot \cite{muhtar2024lhrs}, EarthGPT \cite{zhang2024earthgpt}, EarthMarker \cite{zhang2024earthmarker}, EarthGPT-X \cite{zhang2025earthgpt}, EarthDial \cite{soni2025earthdial}, VHM \cite{pang2025vhm}, GeoPixel \cite{shabbir2025geopixel}, and SkyEyeGPT \cite{zhan2025skyeyegpt}, extend captioning and visual question answering to scene understanding, grounding, multi-sensor dialogue, and pixel-level reasoning~\cite{ye2025whereami,ye2024cross}. These advances strengthen understanding of provided imagery or products but generally do not discover external data, coordinate tool use, or maintain evidence through long-horizon EO analysis.

\textbf{Agent-based EO Methods.} General agents establish planning and tool-use paradigms through reasoning-action interleaving in ReAct \cite{yao2022react}, API-grounded tool use in Gorilla \cite{patil2024gorilla} and ToolLLM \cite{qin2024toolllm}, reusable skills in Voyager \cite{wang2023voyager}, and workflow optimization in AFlow \cite{zhang2025aflow}. Early EO agents, including RS-ChatGPT \cite{guo2024rschatgpt}, RS-Agent \cite{xu2024rs}, and Change-Agent \cite{liu2024change}, connect LLMs with visual models or expert tools for perception tasks. Subsequent systems organize broader workflows through EO code generation in UnivEarth \cite{kao2026towards}, tool use in ThinkGeo \cite{shabbir2025thinkgeo}, MCP-based tool integration in Earth-Agent \cite{feng2025earth}, knowledge-flow fusion in CangLing-KnowFlow \cite{chen2025cangling}, and hierarchical abstraction in HTAM \cite{li2025designing}. Recent directions extend to multi-platform reasoning with RingMo-Agent \cite{hu2025ringmoagent}, climate science with EarthLink \cite{guo2025earthlink}, open geospatial environments with OpenEarthAgent \cite{shabbir2026openearthagent} and OpenEarth-Agent \cite{zhao2026openearth}, workflow automation with GeoFlow \cite{bhattaram2025geoflow}, experience-driven cooperation with GeoEvolver \cite{dai2026geoevolver}, vague-intent handling with RemoteAgent \cite{remoteagent2026}, and expert-level reasoning with GeoMMAgent \cite{xiao2026geommbench}. However, existing systems commonly plan over broad or predefined tool spaces and optimize planning and execution as a unified policy. They rarely use expert task skills to constrain both workflow construction and runtime tool scope, or explicitly maintain accepted evidence and dependencies as execution evolves. Role-specific alignment of workflow composition and node-level argument grounding therefore remains largely underexplored.

\section{Earth-Bench-Pro Benchmark}
\label{sec:earthbenchv2}

\newcommand{\cmark}{\textcolor{green!70!black}{$\checkmark$}}
\newcommand{\xmark}{\textcolor{red}{$\times$}}
\newcommand{\pmark}{\textcolor{black}{Part.}}

\newcolumntype{C}[1]{>{\centering\arraybackslash}m{#1}}

\renewcommand{\tabularxcolumn}[1]{m{#1}}
\newcolumntype{Y}{>{\centering\arraybackslash}X}

\newcommand{\headercell}[1]{%
  \parbox[c][3.8em][c]{\linewidth}{%
    \centering
    \footnotesize
    \bfseries
    #1%
  }%
}

\begin{table*}[t]
  \centering
  \caption{Comparison with recent Earth-observation agent benchmarks.
  Values marked with $^\ast$ are not explicitly reported in the original
  papers and are computed from the corresponding official open-source releases.}
  \label{tab:benchmark_comparison}
  \small
  \setlength{\tabcolsep}{1.8pt}
  \renewcommand{\arraystretch}{1.18}

  \begin{tabularx}{\textwidth}{
    C{2.85cm}
    C{0.75cm}
    C{1.05cm}
    C{0.65cm}
    C{0.70cm}
    *{7}{Y}
  }
    \toprule

    \headercell{Benchmark}
    & \headercell{Tasks}
    & \headercell{Steps}
    & \headercell{Avg.}
    & \headercell{Tools}
    & \headercell{Multi-source\\EO Data}
    & \headercell{Open-ended\\QA}
    & \headercell{Executable\\Trajectory}
    & \headercell{Real Data\\Acquisition}
    & \headercell{Preprocessing\\Required}
    & \headercell{Quantitative\\Answer}
    & \headercell{Intermediate\\Observations} \\

    \midrule

    Ours
    & 744
    & 5,295
    & 7.1
    & 112
    & \cmark
    & \cmark
    & \cmark
    & \cmark
    & \cmark
    & \cmark
    & \cmark \\

    OpenEarth-Bench~\cite{zhao2026openearth}
    & 596
    & --
    & --
    & --
    & \cmark
    & \xmark
    & \cmark
    & \cmark
    & \cmark
    & \cmark
    & \pmark \\

    OpenEarthAgent~\cite{shabbir2026openearthagent}
    & 1,169
    & 7,064
    & 6.04
    & 24
    & \cmark
    & \cmark
    & \cmark
    & \pmark
    & \pmark
    & \cmark
    & \cmark \\

    KnowFlow-Bench~\cite{chen2025cangling}
    & 324
    & --
    & --
    & --
    & \cmark
    & \xmark
    & \cmark
    & \pmark
    & \cmark
    & \cmark
    & \pmark \\

    TerraBench~\cite{nguyen2026terrabench}
    & 403
    & $\sim$24,500
    & $\sim$60.8
    & 77
    & \cmark
    & \cmark
    & \cmark
    & \cmark
    & \pmark
    & \cmark
    & \cmark \\

    EO-Gym~\cite{ma2026eo}
    & 1,436
    & 5,323$^\ast$
    & 3.7$^\ast$
    & 35
    & \cmark
    & \cmark
    & \cmark
    & \pmark
    & \pmark
    & \pmark
    & \cmark \\

    GeoPlan-Bench~\cite{li2025designing}
    & 996
    & 10,619$^\ast$
    & 10.7$^\ast$
    & 104$^\ast$
    & \xmark
    & \xmark
    & \xmark
    & \xmark
    & \cmark
    & \xmark
    & \xmark \\

    \bottomrule
  \end{tabularx}
\end{table*}
\subsection{Benchmark Overview}

Earth-Bench-Pro builds a 744-question EO agent benchmark from 248 matched scientific task cores. Earth-Bench-Pro comprises three matched sub-benchmarks: Earth-Bench-IF, Earth-Bench-AP, and Earth-Bench-OW. Earth-Bench-IF provides an ordered reference tool sequence and prepared inputs. Earth-Bench-AP hides the sequence while retaining prepared inputs. Earth-Bench-OW requires agents to acquire or discover data at runtime and subsequently generate an open-ended answer grounded in execution evidence.

As illustrated in Figure~\ref{benchmark_overview}, Earth-Bench-Pro covers eight application domains and six analysis types. The application domains include agriculture and drought~\cite{li2026can}, urban areas and infrastructure, water and marine systems, vegetation and ecosystems~\cite{zheng2025treecrown}, cryosphere and snow, atmosphere and climate, hazards and disasters, and land cover and object mapping. The analysis types include product statistics, spectral indices, thermal retrieval, brightness-temperature analysis, time-series analysis, and RGB perception. The tasks also span regions across all six inhabited continents and requested observation periods from 1980 to 2025, covering a broad range of geographic settings and temporal extents.

Recent EO agent benchmarks cover only parts of the end-to-end workflow. OpenEarth-Bench~\cite{zhao2026openearth} and KnowFlow-Bench~\cite{chen2025cangling} include acquisition or preprocessing but omit open-ended answers. OpenEarthAgent~\cite{shabbir2026openearthagent}, TerraBench~\cite{nguyen2026terrabench}, and EO-Gym~\cite{ma2026eo} evaluate open-ended answers and tool trajectories but provide partial coverage of acquisition, preprocessing, or quantitative analysis. GeoPlan-Bench~\cite{li2025designing} isolates workflow planning without executable reference trajectories or runtime acquisition. As shown in Table~\ref{tab:benchmark_comparison}, Earth-Bench-Pro is the only benchmark in this comparison that covers all seven dimensions. This design evaluates both final answers and the complete execution processes that produce them.

\subsection{Problem Formulation}

Earth-Bench-Pro represents each benchmark sample as $z=(q,c,\tau^\star,y^\star)$, where $q$ is a scientific query, $c$ encodes the input context or runtime data requirements, and $y^\star$ is the reference answer. Let $N_z$ denote the number of steps in the reference trajectory $\tau^\star$. The EO environment $\mathcal{E}$ executes these steps. Each step records tool $t_i^\star$, arguments $\theta_i^\star$, and the resulting observation $o_i^\star$:
\begin{equation}
    \tau^\star=\bigl((t_i^\star,\theta_i^\star,o_i^\star)\bigr)_{i=1}^{N_z},
    \qquad
    o_i^\star=\mathcal{E}(t_i^\star,\theta_i^\star).
\end{equation}

The three evaluation regimes expose different information to the agent. Let $\pi^\star=(t_1^\star,\ldots,t_{N_z}^\star)$ denote the reference tool sequence, and let $\phi$ construct an instruction that states these tools in order alongside the scientific query. Instruction following uses the resulting instruction $q^{\mathrm{IF}}=\phi(q,\pi^\star)$ and prepared inputs $c^{\mathrm{prep}}$. Autonomous planning uses the query without reference tools and the same prepared inputs. Open-world execution provides runtime input requirements $c^{\mathrm{req}}$, leaving the agent to acquire or discover data during execution:
\begin{equation}
\left\{
\begin{aligned}
    q^{\mathrm{IF}} &= \phi(q,\pi^\star), \\
    \mathcal{I}_{\mathrm{IF}} &= (q^{\mathrm{IF}},c^{\mathrm{prep}}), \\
    \mathcal{I}_{\mathrm{AP}} &= (q,c^{\mathrm{prep}}), \\
    \mathcal{I}_{\mathrm{OW}} &= (q,c^{\mathrm{req}}).
\end{aligned}
\right.
\end{equation}

In the open-world execution, $f$ denotes the tool-selection policy, $g$ denotes the runtime argument generator, and $h$ denotes the evidence-conditioned answer function. The tool and arguments for call $i$ depend on the query, runtime requirements, and preceding accepted execution trace. The environment returns a new observation, and the agent generates its answer from the completed trajectory:
\begin{equation}
\label{eq:open_world_execution}
\left\{
\begin{aligned}
    \tau &= \bigl((t_i,\theta_i,o_i)\bigr)_{i=1}^{L}, \\
    t_i &= f(q,c^{\mathrm{req}},\tau_{<i}), \\
    \theta_i &= g(q,c^{\mathrm{req}},t_i,\tau_{<i}), \\
    o_i &= \mathcal{E}(t_i,\theta_i), \\
    \hat y &= h(q,c^{\mathrm{req}},\tau).
\end{aligned}
\right.
\end{equation}
Execution therefore directly determines the selected data, intermediate dependencies, and ultimately the final answer.

\begin{figure*}[!t]
	\centering
	\includegraphics[width=0.95\textwidth]{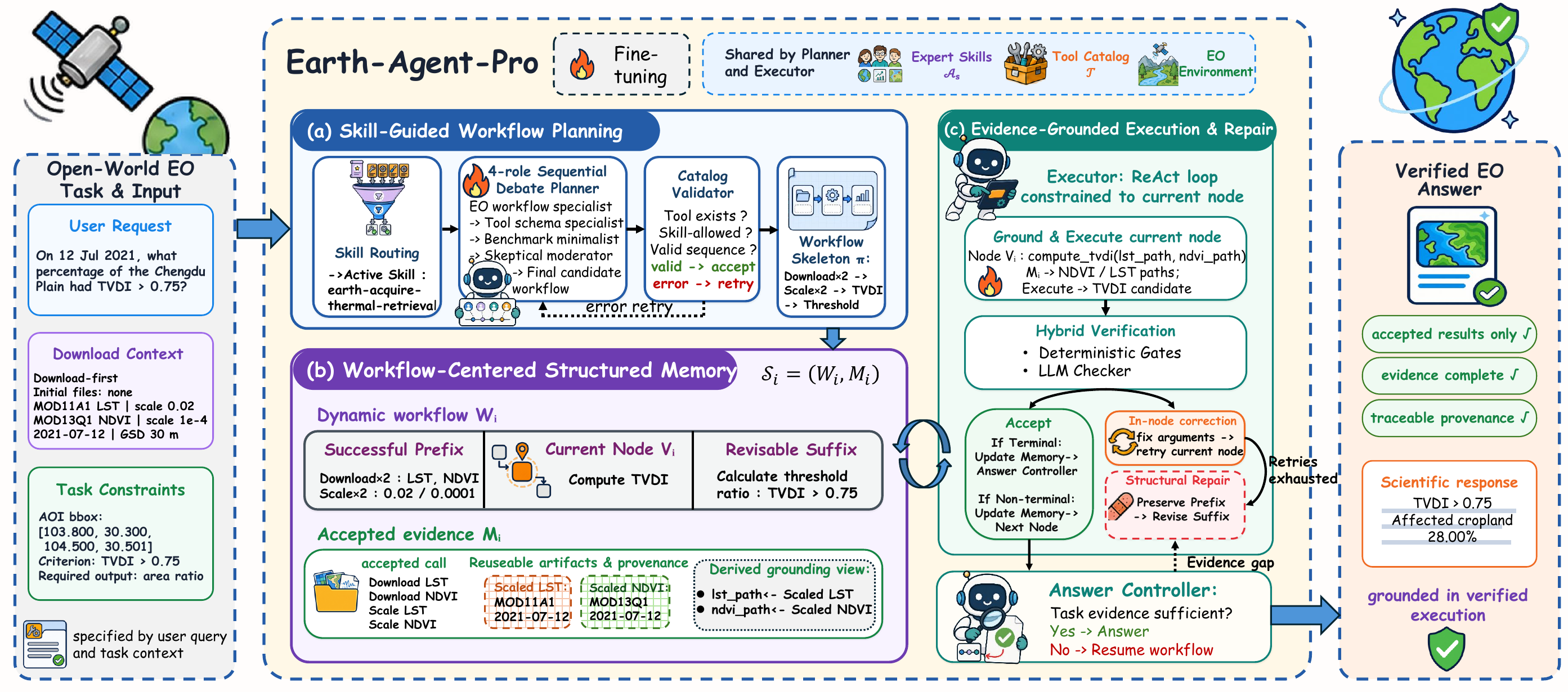}\\
\caption{\textbf{Earth-Agent-Pro Framework.}
\textbf{(a)} Skill-guided workflow planning uses a shared expert specification to constrain both workflow construction and runtime argument grounding.
\textbf{(b)} Workflow-centered structured memory couples the evolving workflow with provenance-aware accepted evidence, preventing failed attempts from contaminating downstream execution.
\textbf{(c)} Evidence-grounded execution and repair performs node-local correction and, upon persistent failure or an evidence gap, preserves the successful prefix while revising only the affected workflow suffix.
Together, these mechanisms enable traceable and bounded adaptation without discarding valid execution progress.}
\label{fig:Earth-Agent-Pro_overview}
\end{figure*}

\subsection{Data Curation}

Earth-Bench-Pro treats a complete executable evidence chain as the unit of annotation, linking each scientific request to its answer through tool execution. We assembled an eight-member annotation team with two computer science experts, three remote sensing specialists, and three Earth science specialists. The team jointly reviews tool executability, remote sensing data requirements, scientific interpretation, and answer correctness, bringing technical constraints and domain judgment into the same annotation process.

Experts transform 248 scientific task cores into open-world workflows for which execution determines the inputs, intermediate dependencies, and answers. These tasks retain the scientific objectives of Earth-Bench~\cite{feng2025earth}. Experts re-execute the original solutions in a shared EO execution environment and correct inconsistencies among the question, tool selection, call arguments, intermediate dependencies, and answer. For 188 Product and Spectrum tasks, experts replace fixed local inputs with online acquisition and required preprocessing, authoring 323 task-level acquisition specifications. Each specification encodes the required source, product, bands, spatial extent, observation period, resolution, sampling interval, and scaling information. The remaining 60 RGB tasks retain user-provided visual inputs, while concrete files still require runtime discovery. After reconstruction, file paths and intermediate artifacts become execution outcomes, and downstream calls must bind the outputs returned by upstream tools.

Each reconstructed task pairs an open-ended query with reference evidence regenerated through execution. Experts remove all answer choices and rewrite each question as a self-contained request specifying the scientific objective, study region, observation period, and task constraints. They rerun the workflow in the same environment, record every call and returned observation, and recompute the answer from the outputs. Each sample retains the representation introduced in Section~\ref{sec:earthbenchv2}:

\begin{equation}
    z=(q,c,\tau^\star,y^\star),\qquad
    \tau^\star=\bigl((t_i^\star,\theta_i^\star,o_i^\star)\bigr)_{i=1}^{N_z}.
\end{equation}

Here, context $c$ records runtime input requirements, and each reference step records tool $t_i^\star$, arguments $\theta_i^\star$, and observation $o_i^\star$. This representation preserves artifact provenance, data dependencies, and answer evidence along the complete execution path.

Quality control checks three relations along the evidence chain: queries against input requirements, data dependencies across tool calls, and final answers against execution evidence. Experts retain and inspect intermediate rasters, files, numerical outputs, detections, masks, and statistics. They also verify answer values, units, precision, temporal references, and trend descriptions. The 248 verified Open-World Execution workflows contain 1,765 tool calls, average 7.1 calls per task, and use 84 distinct tools. We derive matched Instruction-Following and Autonomous-Planning variants from each verified task core, yielding 744 questions and 5,295 reference tool calls. Every answer retains a traceable execution basis, and every intermediate result records its upstream source.

\section{Earth-Agent-Pro: An Execution-Adaptive Agent Framework}
\label{sec:Earth-Agent-Pro}

\subsection{Framework Overview}

Earth-Agent-Pro is a Plan-and-Execute-style agent that separates scientific workflow planning from runtime operation grounding. Before concrete files and intermediate results become available, the Planner constructs an ordered workflow skeleton. As data and intermediate artifacts emerge, the Executor grounds each node, invokes its tool, and updates the state. The dynamic workflow and accepted execution memory form the workflow-centered structured memory shared by both roles throughout the entire execution process. The Planner and workflow controller operationalize the abstract tool-selection policy $f$ in Eq.~\eqref{eq:open_world_execution}, the argument generator $G_{\mathrm{exec}}$ implements $g$, and the Answer Controller implements $h$.

Figure~\ref{fig:Earth-Agent-Pro_overview} presents the complete control process. Let $R_{\mathrm{skill}}$ and $R_{\mathrm{tool}}$ denote skill and tool routing, let $\mathcal T$ denote the global tool catalog, and let $\mathcal A_s$ denote the tools allowed by skill $s$. The two routing stages reduce the catalog to the candidate set $\mathcal P_s$, after which the Planner generates the tool sequence $\pi^{\mathrm{plan}}$ through sequential debate. The initialization operator converts this sequence into dynamic workflow $W_0$ and creates the initial state $\mathcal S_0$ with empty memory $M_0$:
\begin{equation}
\label{eq:agent_initialization}
\left\{
\begin{aligned}
    s &= R_{\mathrm{skill}}(q,c), \\
    \mathcal P_s &= R_{\mathrm{tool}}(q,c,s,\mathcal T\cap\mathcal A_s), \\
    \pi^{\mathrm{plan}} &= \operatorname{Plan}(q,c,s,\mathcal P_s), \\
    \mathcal S_0 &= (W_0,M_0)
    =\bigl(\operatorname{Init}(\pi^{\mathrm{plan}}),\varnothing\bigr).
\end{aligned}
\right.
\end{equation}
The Executor advances node by node over this structured state and exposes only verified results to subsequent operations. When a node fails or the final evidence remains incomplete, the Planner proposes a local replacement suffix and the Executor installs it while retaining the successful prefix.

\subsection{Skill-Guided Workflow Planning}

\begin{figure}[t]
    \centering
    \includegraphics[width=0.95\linewidth]{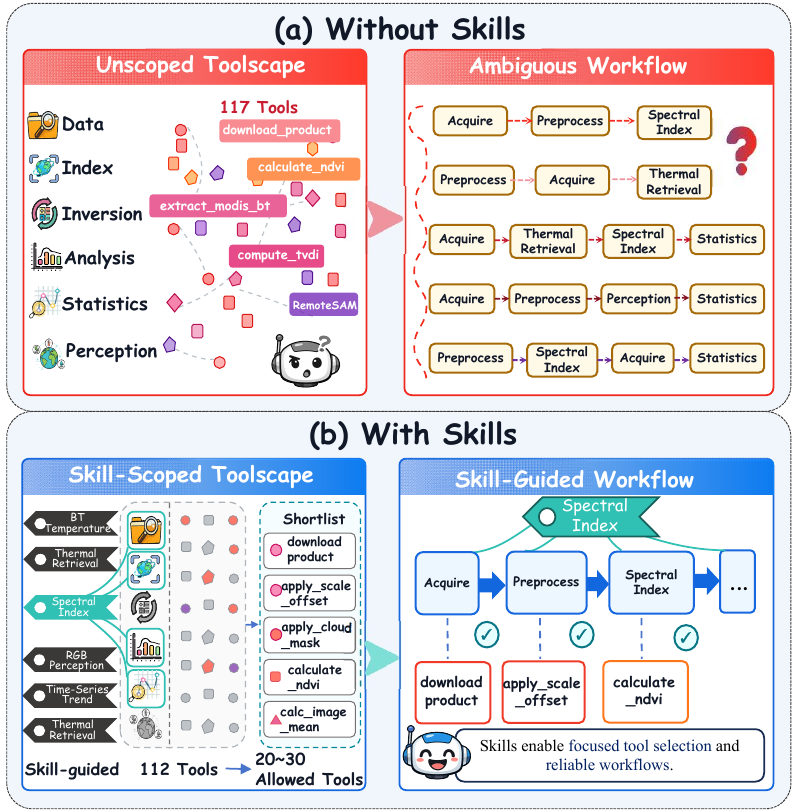}
    \caption{\textbf{With vs. Without Skills.} The top shows an unscoped tool space and ambiguous workflows without skills, while the bottom shows focused tool selection and a coherent workflow enabled by skill guidance.}
    \label{fig:skill_guidance}
\end{figure}

Expert skills encode EO procedure knowledge as task-level specifications shared by the Planner and Executor. We author six skills covering spectral analysis, product-based computation, and RGB perception. Each skill describes its tool scope, operation order, argument constraints, and repetition rules across acquisition, preprocessing, domain computation, and aggregation. As illustrated in Figure~\ref{fig:skill_guidance}, skill guidance removes irrelevant tools before workflow generation and continues to constrain node realization during execution.

The Skill Router and Tool Router implement the two constraints in Eq.~\eqref{eq:agent_initialization}. The full catalog $\mathcal T$ contains 112 callable tools and their schemas. The Skill Router selects a skill from the query and input context, then excludes inapplicable tools through $\mathcal A_s$. The Tool Router retrieves $\mathcal P_s$ for the task objective, leaving the Planner to compose a workflow over legal and task-relevant tools for subsequent execution.

The candidate tools must form a scientifically valid and interface-compatible operation sequence. The Planner uses a four-role sequential debate that examines the EO procedure, tool schemas, sequence parsimony, and final executability in order. Each role revises the candidate sequence using the preceding discussion. A catalog validator then checks tool existence and active-skill membership. It returns validation errors with the candidate sequence for another planning attempt until the Planner obtains $\pi^{\mathrm{plan}}$.

The selected skill remains active after initial planning. For the current operation, the Executor combines the skill instructions with the tool schema, query constraints, and accepted upstream results to generate arguments. When deterministic rules cannot resolve a tool result, the language-model checker uses the same skill to assess whether the result follows the task procedure. The Planner and Executor therefore share one task-level specification for operation selection and runtime argument grounding.

\subsection{Workflow-Centered Structured Memory}

Earth-Agent-Pro represents runtime step $k$ using the structured state $\mathcal S_k=(W_k,M_k)$. The dynamic workflow $W_k$ contains the active node sequence $\mathcal V_k$ and the revision archive $\mathcal R_k$. The execution memory $M_k$ is a chronologically ordered collection of evidence records. Each record $e_u$ stores invoked tool $\bar t_u$, arguments $\theta_u$, observation $o_u$, and provenance $\nu_u$:
\begin{equation}
\label{eq:workflow_state}
\left\{
\begin{aligned}
    \mathcal S_k &= (W_k,M_k), \\
    W_k &= (\mathcal V_k,\mathcal R_k), \\
    \mathcal V_k &= (v_1^k,\ldots,v_{n_k}^k), \\
    M_k &= (e_1,\ldots,e_{m_k}),
    \qquad e_u=(\bar t_u,\theta_u,o_u,\nu_u).
\end{aligned}
\right.
\end{equation}

Each node records its planned tool, invoked tool, execution status, grounded arguments, result summary, failure diagnosis, and revision identifier. Execution memory accumulates only verified calls. Provenance for file artifacts also records variables, time, region, bands, and batches, allowing the Executor to resolve concrete files, numerical series, and cross-input correspondence for the later nodes.

The structured state limits the information exposed to subsequent operations at each step. Argument generation uses the query, current node and schema, active skill, and previously accepted call history. Rejected attempts provide retry feedback for the current node and leave the accepted history unchanged. The Executor therefore avoids repeatedly recovering the valid evidence, failed attempts, and pending operations from an expanding free-form interaction transcript.

\subsection{Evidence-Grounded Execution and Workflow Repair}

The Executor advances the workflow through a ReAct loop constrained to the current node. For node $v_i^k$ in state $\mathcal S_k$, let $t_i^k$ denote its planned tool, let $a$ index in-node attempts, and let $\widetilde t_{i,a}^k$ denote the invoked tool. The first attempt sets $\widetilde t_{i,1}^k=t_i^k$. $\sigma(t)$ returns the schema of tool $t$. Argument generator $G_{\mathrm{exec}}$ combines this schema, the current node, and accepted memory to generate arguments. Execution environment $\mathcal E$ returns an observation, after which hybrid verification gate $V_{\mathrm{hyb}}$ produces decision $\delta_{i,a}^k$ and feedback $b_{i,a}^k$:
\begin{equation}
\label{eq:node_execution}
\left\{
\begin{aligned}
    \hat\theta_{i,a}^k
    &=G_{\mathrm{exec}}(q,c,s,\sigma(\widetilde t_{i,a}^k),v_i^k,W_k,M_k), \\
    \hat o_{i,a}^k
    &=\mathcal E(\widetilde t_{i,a}^k,\hat\theta_{i,a}^k), \\
    (\delta_{i,a}^k,b_{i,a}^k)
    &=V_{\mathrm{hyb}}(q,s,W_k,M_k,\widetilde t_{i,a}^k,
      \hat\theta_{i,a}^k,\hat o_{i,a}^k), \\
    \delta_{i,a}^k
    &\in\{\mathrm{accept},\mathrm{retry},\mathrm{replace}\}.
\end{aligned}
\right.
\end{equation}

The hybrid gate first applies deterministic checks for structural errors, error markers, and batch coverage. The language-model checker evaluates cases that these rules cannot resolve using the query, grounded arguments, active skill, and remaining workflow. Retry updates the arguments and feedback for the current tool. Replace selects another tool from $\mathcal P_s$ and updates the feedback. Both actions remain local to the current attempt and leave persistent state $\mathcal S_k$ unchanged. Accept instead commits the node result to the dynamic workflow and appends its evidence record to execution memory. Let $\nu_{i,a}^k$ denote call provenance, let $U_{\mathrm{acc}}$ denote node commitment, and let $\operatorname{append}$ add an item to an ordered collection. After all in-node attempts fail, the Executor writes the final diagnosis to the failed node and then triggers localized workflow repair:

\begin{equation}
\label{eq:memory_transition}
\begin{aligned}
e_{i,a}^k
&=(\widetilde t_{i,a}^k,\hat\theta_{i,a}^k,
\hat o_{i,a}^k,\nu_{i,a}^k), \\
\mathcal S_{k+1}
&=
\left\{
\begin{array}{@{}ll@{}}
\left(
\begin{aligned}
&U_{\mathrm{acc}}(W_k,i,e_{i,a}^k),\\[-2pt]
&\operatorname{append}(M_k,e_{i,a}^k)
\end{aligned}
\right),

\text{if } \delta_{i,a}^k=\mathrm{accept},
\\[4pt]
\mathcal S_k,

\text{if } \delta_{i,a}^k
\in\{\mathrm{retry},\mathrm{replace}\}.
\end{array}
\right.
\end{aligned}
\end{equation}

The Planner generates candidate suffix $\widetilde{\mathcal V}_{i:}$ from failed node $i$, where $d_i^k$ denotes the final failure diagnosis stored by that node. The Executor validates candidate operations against the active skill and tool catalog. It rejects candidates that repeat an earlier repair, show no execution progress, have abnormal length, or contain excessive operation repetition. After a candidate passes these checks, the Executor uses $\oplus$ to concatenate the successful prefix and new suffix in execution order, then records the old suffix with its diagnosis in $\mathcal R_{k+1}$. The repair leaves $M_k$ unchanged and re-executes only the failed position and subsequent operations:
\begin{equation}
\label{eq:suffix_repair}
\left\{
\begin{aligned}
    &\widetilde{\mathcal V}_{i:}
=\operatorname{Plan}_{\mathrm{repair}}
      (q,c,s,W_k,M_k,i,d_i^k), \\
&\mathcal V_{k+1}
    =\mathcal V_k[1{:}i-1]\mathbin{\oplus}
      \widetilde{\mathcal V}_{i:}, \\
    &\mathcal R_{k+1}
    =\operatorname{append}\bigl(
      \mathcal R_k,(\mathcal V_k[i{:}],\widetilde{\mathcal V}_{i:},d_i^k)\bigr), \\
&W_{k+1}
    =(\mathcal V_{k+1},\mathcal R_{k+1}), \\
&M_{k+1}
=M_k.
\end{aligned}
\right.
\end{equation}

The Answer Controller extends the same repair mechanism to workflows whose tools succeed but whose evidence remains incomplete. It checks required computations, spatiotemporal coverage, aggregation, units, numerical scale, and non-finite results. With sufficient evidence, the controller generates the final answer only from $M_k$. Incomplete evidence produces an actionable diagnosis, after which the Executor reopens the specified node or defaults to the final node when the diagnosis identifies none. Earth-Agent-Pro therefore uses a node as the boundary for runtime correction and a workflow suffix as the boundary for structural revision, respectively.

\section{Role-Specialized Training}
\label{sec:training}

\begin{figure*}[!t]
	\centering
	\includegraphics[width=0.95\textwidth]{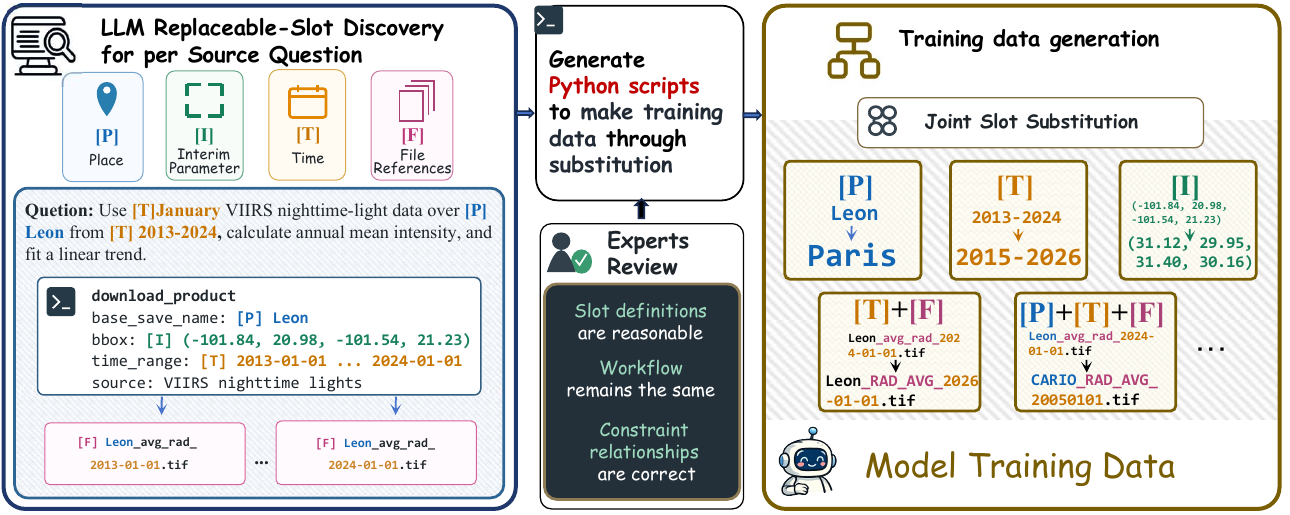}\\
	\caption{\textbf{Synthetic training data generation.} An LLM identifies replaceable slots for each seed sample and writes a sample-specific Python program. Human experts verify that the program preserves workflow structure and dependency constraints before it generates consistent training instances.}\label{fig3}
\end{figure*}
\subsection{Training Overview}
\label{sec:training_overview}

The Planner and Executor expose different prediction interfaces. The Planner composes an ordered tool sequence from expert skill guidance and candidate tool schemas. The Executor grounds a prescribed node from accepted runtime evidence. A monolithic trajectory objective would mix workflow errors with argument errors and spread credit across long tool chains. Role-specialized training isolates these errors at their respective prediction interfaces.

The training pipeline expands instances through dependency-aware trajectory instantiation and then extracts role-specific supervision from complete trajectories. Let $\mathcal D_{\mathrm P}$ and $\mathcal D_{\mathrm E}$ denote the Planner and Executor records. Parameters $\psi_{\mathrm P}$ and $\psi_{\mathrm E}$ are the trainable low-rank adaptation (LoRA) parameters of the two role-specific adapters. Planner context $x^{\mathrm P}$ contains the query, inputs, skill guidance, and candidate tool schemas. Executor context $x_i^{\mathrm E}$ contains the prescribed node and accepted reference-trajectory prefix. For a Planner record, $\boldsymbol\omega^{\mathrm P}$ tokenizes reference tool sequence $\pi^\star$. For Executor sample $j$ at node $i$, $\boldsymbol\omega^{\mathrm E}_{i,j}$ is the response token sequence.

Supervised fine-tuning (SFT) optimizes workflow composition, while on-policy group relative policy optimization (GRPO)~\cite{shao2024deepseekmath} optimizes runtime argument grounding. Let $\hat A_{i,j}$ denote group-relative advantage, and let $\rho_{i,j,\ell}$ and $\bar\rho_{i,j,\ell}$ denote the raw and clipped token-level importance ratios. The frozen reference distribution is $p_{\mathrm{ref}}$. Coefficients $\beta$ and $\eta$ weight the Kullback--Leibler (KL) regularizer and entropy bonus, respectively. The total objective combines the two role losses, and we optimize a separate adapter for each role:
\begin{equation}
\label{eq:role_training_objective}
\left\{
\begin{aligned}
&\mathcal L_{\mathrm{plan}}(\psi_{\mathrm P})
=-\mathbb E_{\mathcal D_{\mathrm P}}
  \sum_\ell \log p_{\psi_{\mathrm P}}
  (\omega_\ell^{\mathrm P}\mid x^{\mathrm P},
  \boldsymbol\omega_{<\ell}^{\mathrm P}), \\
&\mathcal L_{\mathrm{exec}}(\psi_{\mathrm E})
=-\mathbb E_{i,j,\ell}\!\left[
  \min\!\left(\rho_{i,j,\ell}\hat A_{i,j},
  \bar\rho_{i,j,\ell}\hat A_{i,j}\right)\right] \\
&\quad+\beta\,\widehat D_{\mathrm{KL}}
  (p_{\psi_{\mathrm E}}\,\|\,p_{\mathrm{ref}})
  -\eta\,\mathcal H(p_{\psi_{\mathrm E}}), \\
&\mathcal L_{\mathrm{train}}(\psi_{\mathrm P},\psi_{\mathrm E})
=\mathcal L_{\mathrm{plan}}(\psi_{\mathrm P})
  +\mathcal L_{\mathrm{exec}}(\psi_{\mathrm E}), \\
&(\psi_{\mathrm P}^\star,\psi_{\mathrm E}^\star)
\in\underset{\psi_{\mathrm P},\psi_{\mathrm E}}{\arg\min}\,
  \mathcal L_{\mathrm{train}}.
\end{aligned}
\right.
\end{equation}

\subsection{Dependency-Preserving Training Data Construction}
\label{sec:training_data}

Training data follow the executable sample representation of Earth-Bench-Pro. A seed sample is $z=(q,c,\tau^\star,y^\star)$ with $\tau^\star=((t_i^\star,\theta_i^\star,o_i^\star))_{i=1}^{N_z}$. We decompose it into an immutable execution skeleton $\kappa(z)$ and a mutable instance state $\mathbf u(z)$. The skeleton retains the reference tool sequence $\pi^\star$, the corresponding tool schemas, and dependency structure $\mathcal G_z$, a mathematical abstraction of the dependencies encoded by slot groups and cross-node references. The instance state contains locations, spatial extents, temporal ranges, sample identifiers, file references, and replaceable intermediate observations.

For each seed sample, a language model represents the mutable state as typed slots, groups repeated references and dependent fields, and generates a sample-specific instantiation program $F_z$. Given root-slot values $\widetilde{\mathbf u}_{\mathrm{root}}$, the program propagates replacements along these dependencies to produce $\widetilde z$ while preserving $\kappa(\widetilde z)=\kappa(z)$. It therefore instantiates coupled fields such as a place and its bounding box together and propagates changed node outputs to all downstream arguments and observations. Human reviewers inspect the slot specification and program before batch synthesis. Figure~\ref{fig3} shows the process from a seed sample through a reviewed instantiation program to dependency-consistent training instances. Let $R$ denote the number of mutable slots. Operator $\operatorname{Compose}$ reconstructs a sample from its execution skeleton and instance state. We formalize this construction as follows:
\begin{equation}
\label{eq:dependency_preserving_instantiation}
\left\{
\begin{aligned}
&\kappa(z)
=\bigl(\pi^\star,\{\sigma(t_i^\star)\}_{i=1}^{N_z},\mathcal G_z\bigr), \\
&\mathbf u(z)
=(u_1,\ldots,u_R), \\
&z
=\operatorname{Compose}\bigl(\kappa(z),\mathbf u(z)\bigr), \\
&\widetilde z
=F_z\bigl(\kappa(z),\widetilde{\mathbf u}_{\mathrm{root}}\bigr), \\
&\kappa(\widetilde z)
=\kappa(z).
\end{aligned}
\right.
\end{equation}

The reviewed programs fix the sensor or data product, selected bands, ordered tools, argument schemas, processing operations, and dependency structure. They then instantiate samples under a fixed random seed and apply quality filtering. This process yields 2,480 instantiated trajectories derived from 248 seed workflows across RGB imagery, spectral observations, and remote sensing products. We add examples from the OpenEarthAgent training set~\cite{shabbir2026openearthagent} to cover GIS analysis, object-centric visual perception, and geometric reasoning, then carefully balance the two sources to form the final trajectory set $\mathcal Z_{\mathrm{train}}$. The Planner extractor creates one sequence-level record per trajectory with target $\pi^\star$. The Executor extractor creates one argument record per node with target $\theta_i^\star$:
\begin{equation}
\label{eq:role_record_extraction}
\left\{
\begin{aligned}
\mathcal D_{\mathrm P}
&=\{(x^{\mathrm P}(z),\pi^\star(z))
  \mid z\in\mathcal Z_{\mathrm{train}}\}, \\
\mathcal D_{\mathrm E}
&=\{(x_i^{\mathrm E}(z),\theta_i^\star(z))
  \mid z\in\mathcal Z_{\mathrm{train}},\ 1\leq i\leq N_z\}.
\end{aligned}
\right.
\end{equation}

\subsection{Planner SFT for Workflow Composition}
\label{sec:planner_sft}

Planner SFT maps domain-constrained context $x^{\mathrm P}$ to reference workflow $\pi^\star$. In Eq.~\eqref{eq:role_training_objective}, $\mathcal L_{\mathrm{plan}}$ computes autoregressive negative log-likelihood over the assistant response serializing $\pi^\star$. This response contains no tool arguments or observations, so the objective matches the Planner's inference interface.

Each Planner example uses a single-turn conversation format. The system and user messages form $x^{\mathrm P}$ and provide the scientific query, input context, skill guidance, and candidate tools. The assistant target serializes $\pi^\star$ as a JSON object containing only the \texttt{tool\_sequence} field. We mask prompt tokens, so $\mathcal L_{\mathrm{plan}}$ supervises only the assistant response and focuses adaptation on tool selection and ordering.

Parameter-efficient adaptation focuses the Planner's learning signal on its role-specific workflow decisions. We freeze the pretrained backbone throughout the entire optimization process and update only a Planner-specific low-rank adaptation (LoRA) module~\cite{hu2021lora} to learn EO tool-selection and ordering patterns. At inference, the sequential debate and catalog validation described in Section~\ref{sec:Earth-Agent-Pro} enforce procedure, schema, and active-skill constraints beyond the learned workflow prior.

\subsection{Executor GRPO for Runtime Argument Grounding}
\label{sec:executor_rl}

Executor training treats the current workflow node as a local decision interface. Given node context $x_i^{\mathrm E}$ and the schema of prescribed tool $t_i^\star$, policy $p_{\psi_{\mathrm E}}$ generates argument dictionary $\hat\theta_i$ at inference and sampled dictionaries $\hat\theta_{i,j}$ during GRPO. The reward evaluates only argument structure and values because the workflow already fixes the tool identity. This node-level interface isolates feedback for argument grounding from workflow decisions at other nodes.

The Executor reward compares each parsed response directly with the reference argument dictionary $\theta_i^\star$ and requires no learned reward model. Let $\operatorname{parse}$ denote the JSON parser, let $\bot$ denote parse failure, let $\mathbb J$ denote the set of valid JSON values, and let $\mathbb D_+\subset\mathbb J$ denote the set of nonempty dictionaries. For candidate $j$ at node $i$, $\boldsymbol\omega_{i,j}^{\mathrm E}$ denotes the raw token sequence, $\xi_{i,j}=\operatorname{parse}(\boldsymbol\omega_{i,j}^{\mathrm E})$ denotes its parsed value, and we set $\hat\theta_{i,j}=\xi_{i,j}$ when $\xi_{i,j}\in\mathbb D_+$. For a nonempty dictionary, $K^\star$ and $\hat K$ denote the reference and predicted key sets, $K_c=K^\star\cap\hat K$ and $K_d=K^\star\triangle\hat K$ denote the common and discrepant keys, and $m=|K^\star|>0$. Value-agreement term $A_{\mathrm{val}}$ counts type-aware matches over common keys under relation $\equiv$, which recursively compares nested values and their types with a tolerance of $10^{-6}$ for floating-point values. The following equation combines key coverage, a symmetric-difference penalty, and value agreement in $S_{\mathrm{arg}}$, then defines complete reward $r_{i,j}$. A missing key incurs a larger penalty than a superfluous key because it reduces coverage and increases the symmetric difference. Runtime verification separately determines execution validity:
\begin{equation}
\label{eq:executor_argument_score}
\begin{aligned}
A_{\mathrm{val}}
&=\sum_{k\in K_c}
  \mathbb I[\hat\theta_{i,j,k}\equiv\theta_{i,k}^\star], \\
S_{\mathrm{arg}}(\hat\theta_{i,j},\theta_i^\star)
&=\max\!\left\{-1,
  \frac{|K_c|-|K_d|+A_{\mathrm{val}}}{m}\right\}, \\
r_{i,j}
&=
\begin{cases}
-2, & \xi_{i,j}=\bot, \\
-1, & \xi_{i,j}\in\mathbb J\setminus\mathbb D_+, \\
S_{\mathrm{arg}}(\hat\theta_{i,j},\theta_i^\star),
    & \xi_{i,j}\in\mathbb D_+.
\end{cases}
\end{aligned}
\end{equation}

For each node input, we sample $G=4$ candidate argument dictionaries from the current policy and standardize their rewards within the group. Let $r_{i,j}$ denote the reward for candidate $j$, let $p_{\mathrm{old}}$ denote the policy that generated it, and let $\epsilon_c$ denote the GRPO clipping radius. Relative advantage $\hat A_{i,j}$ supplies a quality signal under the same node context without a separate critic. In Eq.~\eqref{eq:role_training_objective}, the KL regularizer limits deviation from the reference model, and the entropy bonus discourages premature collapse. During training, we freeze the pretrained backbone and update only the Executor-specific LoRA adapter. The trained policy grounds arguments for the current node, while explicit runtime mechanisms control environment interaction, observation verification, retries, and workflow repair. The group-relative advantage, token-level importance ratio, and its clipped form define the GRPO surrogate term:
\begin{equation}
\label{eq:executor_grpo_terms}
\left\{
\begin{aligned}
\hat A_{i,j}
&=\frac{r_{i,j}-\operatorname{mean}_{j'}(r_{i,j'})}
        {\operatorname{std}_{j'}(r_{i,j'})+10^{-6}}, \\
\rho_{i,j,\ell}
&=\frac{p_{\psi_{\mathrm E}}
  (\omega_{i,j,\ell}^{\mathrm E}\mid x_i^{\mathrm E},
   \boldsymbol\omega_{i,j,<\ell}^{\mathrm E})}
  {p_{\mathrm{old}}
  (\omega_{i,j,\ell}^{\mathrm E}\mid x_i^{\mathrm E},
   \boldsymbol\omega_{i,j,<\ell}^{\mathrm E})}, \\
\bar\rho_{i,j,\ell}
&=\operatorname{clip}_{[1-\epsilon_c,\,1+\epsilon_c]}
  (\rho_{i,j,\ell}).
\end{aligned}
\right.
\end{equation}

\section{Experiments}
\label{sec:experiments}

\begin{table*}[t]
  \centering
  \caption{Performance of different LLM backbones with Earth-Agent-Pro on
  Earth-Bench-Pro. All quality metrics are percentages. Efficiency is a ratio
  for which values closer to $1$ are better. Among open-source models, best results are bolded and
  runner-ups are underlined.}
  \label{tab:exp-v2-backbones}
  \small
  \setlength{\tabcolsep}{2.8pt}
  \renewcommand{\arraystretch}{1.18}

  \begin{tabularx}{0.95\textwidth}{
    L{5.0cm}
    *{8}{Y}
  }
    \toprule

    \textbf{Model}
    & \makecell[c]{\textbf{LLM-as-}\\\textbf{Judge}}
    & \textbf{ROUGE-L}
    & \textbf{EM}
    & \textbf{Efficiency}
    & \textbf{TAO}
    & \textbf{TIO}
    & \textbf{TEM}
    & \textbf{Param.} \\

    \midrule
    \rowcolor{gray!14}
    GPT-5
      & 66.13
      & 21.68
      & 13.31
      & 1.0911
      & 87.61
      & 71.97
      & 60.14
      & 27.28 \\

    \rowcolor{gray!14}
    Gemini-3.5-flash
      & 63.31
      & 25.43
      & 20.56
      & 1.0752
      & 89.28
      & 73.90
      & 64.49
      & 32.19 \\

    \rowcolor{gray!14}
    Claude-4.6-sonnet
      & 56.85
      & 29.09
      & 29.03
      & 1.1911
      & 90.17
      & 76.59
      & 63.52
      & 28.02 \\

    \midrule
    GLM-4.5-Air
      & 48.39
      & 20.45
      & \underline{20.16}
      & 1.0063
      & 85.46
      & 72.76
      & 62.30
      & 27.01 \\

    Llama-3.1-8B-Instruct
      & 13.71
      & 5.15
      & 2.02
      & 0.9811
      & 75.34
      & 60.81
      & 44.95
      & 22.04 \\

    Llama-3.2-3B-Instruct
      & 4.44
      & 2.25
      & 2.42
      & 0.6873
      & 51.09
      & 44.26
      & 36.62
      & 13.96 \\

    Llama-3.3-70B-Instruct
      & 33.87
      & 8.69
      & 6.45
      & 1.0526
      & 85.17
      & 73.51
      & 56.69
      & 27.39 \\

    Llama-4-Scout
      & 32.66
      & 9.91
      & 7.66
      & 0.9335
      & 81.81
      & 70.47
      & 57.57
      & 25.88 \\

    InternLM3-8B
      & 19.35
      & 6.86
      & 5.24
      & 0.7617
      & 69.33
      & 58.49
      & 48.85
      & 20.73 \\

    Intern-S1
      & 40.73
      & 15.49
      & 13.71
      & \underline{1.0022}
      & 85.44
      & 73.62
      & 60.24
      & \underline{31.26} \\

    GPT-oss-20B
      & 38.71
      & 14.10
      & 10.48
      & 1.1510
      & 85.94
      & 71.08
      & 59.59
      & 28.62 \\

    GPT-oss-120B
      & 48.39
      & 13.54
      & 8.87
      & 1.0415
      & 86.16
      & 71.62
      & 61.61
      & 29.84 \\

    Qwen3.5-27B
      & \underline{49.60}
      & 19.12
      & 15.32
      & \textbf{1.0017}
      & 87.67
      & 75.27
      & 68.71
      & 29.86 \\

    Gemma-4 (E4B-it)
      & 31.45
      & 12.49
      & 9.68
      & 0.7841
      & 69.33
      & 61.36
      & 55.13
      & 23.26 \\

    Gemma-4 (26B-A4B-it)
      & 47.58
      & 18.49
      & 14.52
      & 1.0766
      & 84.19
      & 72.62
      & 61.80
      & 31.23 \\

    MistralAI-v0.3-7B
      & 14.11
      & 6.32
      & 4.44
      & 0.5876
      & 54.31
      & 46.18
      & 42.18
      & 15.61 \\

    \midrule

    \addlinespace[1.5pt]

    Qwen3.5-4B
      & 32.26
      & 16.61
      & 14.52
      & 0.9447
      & 79.21
      & 66.87
      & 57.99
      & 21.58 \\

    Qwen3.5-9B
      & 38.31
      & 16.33
      & 15.73
      & 1.0129
      & 85.59
      & 74.57
      & 67.29
      & 25.05 \\

    \textbf{Ours (4B, fine-tuned)}
      & 43.15
      & \textbf{22.42}
      & \textbf{21.77}
      & 1.1117
      & \underline{87.93}
      & \underline{77.52}
      & \underline{68.96}
      & 28.98 \\

    \textbf{Ours (9B, fine-tuned)}
      & \textbf{50.00}
      & \underline{20.65}
      & 19.76
      & 1.1571
      & \textbf{91.28}
      & \textbf{81.11}
      & \textbf{71.01}
      & \textbf{32.77} \\
    \bottomrule
  \end{tabularx}
\end{table*}
We evaluate Earth-Agent-Pro from seven perspectives. We first examine its sensitivity to LLM backbones and compare it with remote-sensing MLLMs. We then assess cross-benchmark transfer on
the OpenEarthAgent benchmark~\cite{shabbir2026openearthagent} and compare Earth-Agent-Pro with general-purpose agents. Next, we
isolate the effect of the agent framework through controlled comparisons on
Earth-Bench-Pro. Finally, we ablate domain skills and disentangle the
contributions of Planner SFT and Executor GRPO~\cite{shao2024deepseekmath} through adapter composition,
planning-only evaluation, and oracle-plan execution. Together, these
experiments evaluate the effectiveness and transferability of Earth-Agent-Pro
and identify the mechanisms.

\subsection{Experimental Setup}

\paragraph{Benchmarks and Comparison Groups}
Our main evaluation is conducted on Earth-Bench-Pro
(Section~\ref{sec:earthbenchv2}), which converts Earth-Bench into open-ended,
end-to-end Earth-observation tasks that require data acquisition,
preprocessing, analysis, and evidence-grounded answer generation. We
additionally evaluate transfer to the OpenEarthAgent benchmark~\cite{shabbir2026openearthagent},
compare against remote-sensing MLLMs on the categories used by
TerraScope~\cite{shu2026terrascope}, and compare with general-purpose agents
across Spectrum, Products, and RGB tasks.

\paragraph{Baselines and Models}

We first evaluate Earth-Agent-Pro with three closed-weight LLM backbones,
GPT-5, Gemini-3.5-flash, and Claude-4.6-sonnet, and open-weight backbones from
the GLM, Llama, InternLM, GPT-oss, Qwen, Gemma, and Mistral families. We also
include fine-tuned Qwen3.5-9B and Qwen3.5-4B variants. We conduct
this training and all open-weight model evaluations on eight NVIDIA H200
graphics processing units (GPUs), use vLLM for model serving, and query the
three closed-weight models through their official APIs. The remote-sensing
comparison includes GeoChat~\cite{kuckreja2024geochat},
TeoChat~\cite{irvin2025teochat}, EarthDial~\cite{soni2025earthdial},
LHRS-Bot~\cite{muhtar2024lhrs}, EarthMind~\cite{shu2025earthmind}, and
TerraScope~\cite{shu2026terrascope}. For framework-level evaluation on
the OpenEarthAgent benchmark~\cite{shabbir2026openearthagent}, we compare ReAct~\cite{yao2022react},
AFlow~\cite{zhang2025aflow}, OpenEarthAgent~\cite{shabbir2026openearthagent}, and
Earth-Agent-Pro under a shared GPT-5 backbone. We next compare against the
general-purpose agents Codex and Manus. Finally, we repeat the framework
comparison on Earth-Bench-Pro using the same GPT-5 backbone to more clearly
isolate the specific effect of the agent framework itself.

\paragraph{Metrics}
We adopt benchmark-specific evaluation protocols. On Earth-Bench-Pro, we
mainly follow the dual-level evaluation protocol of
Earth-Agent~\cite{feng2025earth}. Final-answer quality is evaluated using
LLM-as-Judge accuracy, ROUGE-L~\cite{lin2004rouge}, and embedding-based semantic matching (EM),
which respectively measure judged correctness, lexical overlap, and semantic
similarity to the reference answer. Execution trajectories are evaluated
using Efficiency, Tools-Any-Order (TAO), Tools-In-Order (TIO),
Tool-Exact-Match (TEM), and Parameter Accuracy. These trajectory metrics collectively characterize tool coverage, tool ordering, strict agreement with the reference workflow, and the correctness of tool arguments.

For the OpenEarthAgent benchmark~\cite{shabbir2026openearthagent}, we retain its official evaluation
protocol, including category-wise tool-selection F1, tool-order fidelity, and
task-level accuracy. For the final-result comparisons in
Tables~\ref{tab:exp-rsmllm} and~\ref{tab:exp-general-agents}, we report
accuracy only. A non-numerical prediction is counted as correct when it
matches the reference answer, while a numerical prediction is considered correct when its relative error with respect to the ground truth is below 15\%. All reported accuracy values are computed as the percentage of correct predictions over the evaluated samples for each corresponding comparison.

\subsection{Main Results on Earth-Bench-Pro}

\paragraph{Backbone and Fine-Tuning Results}

We first examine how the policy backbone affects Earth-Agent-Pro on the open-ended Earth-Bench-Pro tasks. We take the off-the-shelf Qwen3.5-4B and Qwen3.5-9B models as our baselines, enabling direct comparison with our fine-tuned variants at the same model scales. Table~\ref{tab:exp-v2-backbones} reports final answer quality, trajectory length, and step-level agreement with expert workflows under the same unified evaluation protocol.

\textbf{Observation 1: answer and trajectory metrics reveal different model
strengths.}
GPT-5 obtains the highest LLM-as-Judge score (66.13\%), whereas the fine-tuned
4B model leads on ROUGE-L and EM. In contrast, the fine-tuned 9B model achieves
the strongest TAO, TIO, TEM, and parameter accuracy, showing that final-answer
quality and workflow fidelity do not necessarily improve together.

\textbf{Observation 2: role-specialized training consistently strengthens
small backbones.}
Relative to Qwen3.5-9B, the fine-tuned 9B variant improves LLM-as-Judge accuracy from
38.31\% to 50.00\% and raises parameter accuracy by 7.72 points. The 4B model
exhibits similarly broad gains, including increases of 10.89 points in Judge
accuracy and 10.97 points in TEM. These results indicate that the training
procedure improves both answer generation and structured tool execution.

\subsection{Comparisons with Existing Methods and Agents}

\paragraph{Remote-Sensing MLLM Comparison}

\newcolumntype{L}[1]{>{\raggedright\arraybackslash}m{#1}}
\newcolumntype{C}[1]{>{\centering\arraybackslash}m{#1}}

\begin{table*}[t]
  \centering
  \caption{Comparison with general RS MLLMs across the TerraScope task categories. A numerical prediction is counted as correct if its relative error from the ground-truth value is less than 15\%. The Avg. column reports sample-weighted accuracy across the five task categories.
  Best results are in bold, and second-best are underlined.}
  \label{tab:exp-rsmllm}

  \small
  \setlength{\tabcolsep}{3.5pt}
  \renewcommand{\arraystretch}{1.22}

  \begin{tabular}{
    L{3.95cm}
    C{2.15cm}
    C{2.15cm}
    C{2.15cm}
    C{2.15cm}
    C{2.15cm}
    C{1.05cm}
  }
    \toprule

    \textbf{Model}
    & \makecell[c]{\textbf{Boundary}\\\textbf{Detection}\\\textbf{(855)}}
    & \makecell[c]{\textbf{Comparative}\\\textbf{Ranking}\\\textbf{(855)}}
    & \makecell[c]{\textbf{Coverage}\\\textbf{Estimation}\\\textbf{(855)}}
    & \makecell[c]{\textbf{Area}\\\textbf{Measurement}\\\textbf{(855)}}
    & \makecell[c]{\textbf{Distance}\\\textbf{Measurement}\\\textbf{(129)}}
    & \textbf{Avg.} \\

    \midrule

    GeoChat~\cite{kuckreja2024geochat}
      & 57.54
      & 58.83
      & 12.16
      & 0.12
      & \underline{0.78}
      & 31.02 \\

    TeoChat~\cite{irvin2025teochat}
      & 56.14
      & 65.85
      & 7.25
      & 0.12
      & 0.00
      & 31.16 \\

    EarthDial~\cite{soni2025earthdial}
      & 55.09
      & 52.98
      & 7.02
      & 0.00
      & 0.00
      & 27.73 \\

    LHRS-Bot~\cite{muhtar2024lhrs}
      & 55.20
      & 56.73
      & 8.07
      & 1.55
      & 0.00
      & 28.97 \\

    EarthMind~\cite{shu2025earthmind}
      & 54.85
      & 59.77
      & 12.28
      & \underline{4.65}
      & 0.00
      & 30.74 \\

    TerraScope~\cite{shu2026terrascope}
      & \underline{66.20}
      & 69.94
      & \textbf{40.70}
      & 0.00
      & 0.00
      & 42.60 \\

    \addlinespace[2pt]
    \midrule
    Earth-Agent-Pro (GPT-5)
      & \textbf{66.43}
      & \underline{70.64}
      & \underline{40.35}
      & \textbf{31.58}
      & \textbf{1.55}
      & \underline{50.41} \\

    Earth-Agent-Pro (Qwen3.5-9B)
      & \textbf{66.43}
      & \textbf{71.35}
      & \underline{40.35}
      & \textbf{31.58}
      & \textbf{1.55}
      & \textbf{50.58} \\

    \bottomrule
  \end{tabular}
\end{table*}

\begin{table*}[t]
  \centering
  \caption{Framework Comparison on the OpenEarthAgent Benchmark. We report tool-selection
  F1 scores for Perception (Per.), Operation (Op.), Logic, and GIS;
  tool-order fidelity under AnyOrder, SameOrder, and Unique; and task-level
  accuracy for nongenerative answering (Ans.) and image generation (Gen.). Best results are in bold, and second-best results are underlined.}
  \label{tab:exp-openearth}

  \small
  \setlength{\tabcolsep}{2.4pt}
  \renewcommand{\arraystretch}{1.20}

  \begin{tabularx}{0.95\textwidth}{
    L{4.15cm}
    *{9}{Y}
  }
    \toprule

    \multirow{2}{*}{\textbf{Framework}}
    & \multicolumn{4}{c}{\textbf{Tool Selection F1}}
    & \multicolumn{3}{c}{\textbf{Tool-Order Fidelity}}
    & \multicolumn{2}{c}{\textbf{Task Accuracy}} \\

    \cmidrule(lr){2-5}
    \cmidrule(lr){6-8}
    \cmidrule(lr){9-10}

    & \textbf{Per.}
    & \textbf{Op.}
    & \textbf{Logic}
    & \textbf{GIS}
    & \textbf{AnyOr.}
    & \textbf{SameOr.}
    & \textbf{Uni.}
    & \textbf{Ans.}
    & \textbf{Gen.} \\

    \midrule

    ReAct (GPT-5)
      & 40.87
      & \underline{66.90}
      & 17.18
      & 88.41
      & 46.71
      & 46.71
      & 48.59
      & 39.45
      & \textbf{80.69} \\

    AFlow (GPT-5)
      & 32.54
      & \textbf{67.69}
      & 36.30
      & 86.75
      & 51.11
      & 50.56
      & 54.17
      & 45.38
      & 65.12 \\

    \addlinespace[1.5pt]

    OpenEarthAgent (GPT-5)
      & 16.88
      & 47.00
      & 6.26
      & 91.50
      & 46.96
      & 46.79
      & 47.81
      & 43.88
      & 46.21 \\

    OpenEarthAgent-4B
      & 57.49
      & 49.12
      & \textbf{44.80}
      & 94.51
      & 61.50
      & 60.65
      & 67.32
      & 43.25
      & 71.55 \\

    Earth-Agent-Pro (GPT-5)
      & \textbf{62.30}
      & 55.68
      & 35.15
      & \underline{95.59}
      & 64.23
      & 64.80
      & 70.97
      & \textbf{56.54}
      & \underline{74.01} \\

    \midrule

    \textbf{Ours (4B, fine-tuned)}
      & 56.16
      & 63.23
      & \underline{42.92}
      & 95.12
      & \underline{70.13}
      & \textbf{69.77}
      & \textbf{75.87}
      & 46.86
      & 72.26 \\
    \textbf{Ours (9B, fine-tuned)}
      & \underline{57.70}
      & 44.79
      & \textbf{44.80}
      & \textbf{96.12}
      & \textbf{70.58}
      & \underline{68.70}
      & \underline{73.95}
      & \underline{47.06}
      & 72.99 \\

    \bottomrule
  \end{tabularx}
\end{table*}

Table~\ref{tab:exp-rsmllm} reports the performance of Earth-Agent-Pro and representative remote-sensing MLLMs on five TerraScope task categories, including boundary detection, comparative ranking, coverage estimation, area measurement, and distance measurement. The evaluation uses open-ended questions instead of multiple-choice options, requiring models to directly produce semantic or numerical answers without candidate guidance.

\textbf{Observation 1. Remote-sensing MLLMs  remain weak on quantitative reasoning.}
Across boundary detection and comparative ranking, existing remote-sensing MLLMs achieve moderate accuracy, which suggests that these models can often recognize coarse land-cover semantics and compare visible regions. However, their performance drops sharply on coverage estimation, area measurement, and distance measurement, indicating that remote-sensing MLLMs still lack reliable metric grounding when exact numerical answers are required.

\textbf{Observation 2. Earth-Agent-Pro achieves the highest overall accuracy.}
Earth-Agent-Pro obtains the best average accuracy among compared methods,
reaching 50.58\% with Qwen3.5-9B and 50.41\% with GPT-5. Compared with the
strongest baseline, TerraScope (42.60\%), these results correspond to
improvements of 7.98 and 7.81 percentage points, respectively. The advantage
is most pronounced in area measurement, where both Earth-Agent-Pro variants
achieve 31.58\%, compared with at most 4.65\% among the remote-sensing MLLMs.
Earth-Agent-Pro maintains strong semantic performance, achieving 66.43\%
on boundary detection and up to 71.35\% on comparative ranking. These results
demonstrate its effectiveness across both semantic understanding and
quantitative reasoning tasks in open-ended evaluation settings.

\paragraph{Cross-Benchmark Comparison on OpenEarthAgent}

We further evaluate the cross-benchmark transferability of Earth-Agent-Pro on
the OpenEarthAgent benchmark~\cite{shabbir2026openearthagent}. To isolate the effect of the agent
framework, we first compare ReAct, AFlow, OpenEarthAgent, and Earth-Agent-Pro using the
same GPT-5 backbone and the official OpenEarthAgent benchmark~\cite{shabbir2026openearthagent} evaluation protocol. We
additionally evaluate the role-specialized 9B and 4B variants to examine
whether backbones can retain competitive workflow-planning ability.

\textbf{Observation 1: Earth-Agent-Pro improves cross-benchmark workflow fidelity.}
Under the shared GPT-5 backbone and evaluation protocol, Earth-Agent-Pro achieves
the strongest Perception, GIS, and all three tool-order scores. It also
improves answering accuracy by 11.16 points over the strongest GPT-5 baseline,
indicating that the framework advantage transfers beyond Earth-Bench-Pro.

\textbf{Observation 2: compact fine-tuned models retain strong procedural
capabilities.}
The fine-tuned 9B model leads on GIS and AnyOrder, while the 4B model achieves
the best SameOrder and Unique scores. These results suggest that
role-specialized training enables small backbones to learn transferable
workflow patterns across distinct benchmark settings.

\begin{table*}[t]
  \centering
  \caption{Comparison with general-purpose agents. Accuracy values are
  reported as percentages, and average latency is measured in seconds.
  Best results are bolded and runner-ups are underlined.}
  \label{tab:exp-general-agents}

  \small
  \setlength{\tabcolsep}{4.0pt}
  \renewcommand{\arraystretch}{1.22}

  \begin{tabularx}{0.95\textwidth}{
    L{6.0cm}
    *{5}{Y}
  }
    \toprule

    \multirow{2}{*}{\textbf{Framework}}
    & \multicolumn{4}{c}{\textbf{Accuracy (\%)}}
    & \multirow{2}{*}{\makecell[c]{\textbf{Avg. Latency}\\\textbf{(s)}}} \\

    \cmidrule(lr){2-5}

    & \textbf{Spectrum}
    & \textbf{Products}
    & \textbf{RGB}
    & \textbf{Avg.}
    & \\

    \midrule

    Codex
      & 40.00
      & \textbf{75.00}
      & 25.00
      & 46.67
      & 268.73 \\

    Manus
      & 40.00
      & 45.00
      & 45.00
      & 43.33
      & 291.00 \\

    \midrule

    Earth-Agent-Pro (GPT-5)
      & \textbf{70.00}
      & \underline{55.00}
      & \textbf{100.00}
      & \textbf{75.00}
      & 208.64 \\

    Earth-Agent-Pro (Qwen3.5-9B)
      & \underline{45.00}
      & 40.00
      & \underline{65.00}
      & 50.00
      & \textbf{67.83} \\

    \textbf{Earth-Agent-Pro (Qwen3.5-9B, fine-tuned)}
      & \textbf{70.00}
      & 40.00
      & \underline{65.00}
      & \underline{58.33}
      & \underline{111.79} \\

    \bottomrule
  \end{tabularx}
\end{table*}

\begin{table*}[t]
  \centering
  \caption{Framework comparison on Earth-Bench-Pro using a shared GPT-5
  backbone. All quality metrics are reported as percentages. Efficiency is
  measured as a ratio, where values closer to $1$ indicate better performance.
  Best results are bolded and runner-ups are underlined.}
  \label{tab:exp-v2-frameworks}

  \small
  \setlength{\tabcolsep}{3.6pt}
  \renewcommand{\arraystretch}{1.22}

  \begin{tabularx}{0.95\textwidth}{
    L{3.8cm}
    *{8}{Y}
  }
    \toprule

    \textbf{Framework}
      & \makecell[c]{\textbf{LLM-as-}\\\textbf{Judge}}
      & \mbox{\textbf{ROUGE-L}}
      & \textbf{EM}
      & \textbf{Efficiency}
      & \textbf{TAO}
      & \textbf{TIO}
      & \textbf{TEM}
      & \textbf{Param.} \\

    \midrule

    ReAct
      & \underline{45.18}
      & \underline{19.75}
      & \underline{26.60}
      & \underline{0.8636}
      & 57.83
      & \underline{47.53}
      & \underline{43.33}
      & \underline{24.33} \\
    
    AFlow
      & 40.16
      & 10.80
      & \textbf{36.70}
      & 0.8511
      & \underline{70.47}
      & 45.46
      & 37.18
      & 20.33 \\

    OpenEarthAgent
      & 41.53
      & 13.83
      & 25.00
      & 0.8101
      & 60.06
      & 42.62
      & 39.32
      & 22.67 \\

    \textbf{Earth-Agent-Pro}
      & \textbf{66.13}
      & \textbf{21.68}
      & 13.31
      & \textbf{1.0911}
      & \textbf{87.61}
      & \textbf{71.97}
      & \textbf{60.14}
      & \textbf{27.28} \\

    \bottomrule
  \end{tabularx}
\end{table*}
\paragraph{General-Purpose Agent Comparison}

We compare Earth-Agent-Pro with Codex and Manus across Spectrum, Products, and RGB
tasks. Table~\ref{tab:exp-general-agents} reports both task accuracy and mean
wall-clock latency, exposing the trade-off between accuracy and execution cost.

\textbf{Observation 1: a domain-specific workflow improves
overall completion.}
Earth-Agent-Pro (GPT-5) obtains 75.00\% average accuracy, 28.33 points above
Codex, while also reducing average latency from 268.73\,s to 208.64\,s. Its
largest advantage appears on RGB tasks, where it completes all evaluated
examples. The simultaneous gains in accuracy and latency indicate that
improved task completion does not necessarily require higher execution cost.

\paragraph{Framework Comparison on Earth-Bench-Pro}

We now directly isolate the effect of the Earth-Agent-Pro framework in a strictly controlled setting by comparing it with ReAct, AFlow, and OpenEarthAgent under the same GPT-5 backbone. Table~\ref{tab:exp-v2-frameworks} provides evidence for the framework's effectiveness on our target benchmark.

\textbf{Observation 1: Earth-Agent-Pro leads seven of eight framework-comparison
metrics.}
It improves LLM-as-Judge accuracy by 20.95 points over the strongest baseline and
raises TAO from 70.47\% to 87.61\%. The corresponding gains in TIO, TEM, and
parameter accuracy show that the improvement extends beyond selecting a
plausible tool set to preserving more of the expert workflow structure.

\subsection{Ablation Studies}

\paragraph{Domain-Skill Ablation}

We remove the retrieved domain skill while keeping the framework and
model fixed within each pair. Table~\ref{tab:exp-skill-ablation} reports the
result for ReAct, Earth-Agent-Pro with the base 9B model, and Earth-Agent-Pro with the
fine-tuned 9B model. Bold values indicate the better result within each
with/without-skill pair; for Efficiency, better means closer to $1$.

\begin{table*}[t]
  \centering
  \caption{Ablation study of domain skills. All quality metrics are reported
  as percentages, while Efficiency is measured as a ratio. Bold values
  indicate the better result within each adjacent pair.}
  \label{tab:exp-skill-ablation}

  \small
  \setlength{\tabcolsep}{2.3pt}
  \renewcommand{\arraystretch}{1.20}

  \begin{tabularx}{0.95\textwidth}{
    L{2cm}
    L{2cm}
    C{1.2cm}
    *{8}{Y}
  }
    \toprule

    \textbf{Framework}
      & \textbf{Model}
      & \textbf{Skills}
      & \makecell[c]{\textbf{LLM-as-}\\\textbf{Judge}}
      & \mbox{\textbf{ROUGE-L}}
      & \textbf{EM}
      & \textbf{Efficiency}
      & \textbf{TAO}
      & \textbf{TIO}
      & \textbf{TEM}
      & \textbf{Param.} \\

    \midrule

    \multirow{2}{*}{ReAct}
      & \multirow{2}{*}{Qwen3.5-9B}
      & $\times$
      & 34.27
      & 13.97
      & 10.08
      & \textbf{2.1549}
      & 56.46
      & 31.39
      & 21.55
      & 10.44 \\

      &
      & $\checkmark$
      & \textbf{39.92}
      & \textbf{14.00}
      & \textbf{11.69}
      & 2.2145
      & \textbf{57.13}
      & \textbf{37.66}
      & \textbf{29.76}
      & \textbf{12.14} \\

    \midrule

    \multirow{4}{*}{Earth-Agent-Pro}
      & \multirow{2}{*}{Qwen3.5-9B}
      & $\times$
      & 35.08
      & 15.99
      & 15.32
      & \textbf{0.9955}
      & 75.61
      & 61.94
      & 51.71
      & 24.84 \\

      &
      & $\checkmark$
      & \textbf{38.31}
      & \textbf{16.33}
      & \textbf{15.73}
      & 1.0129
      & \textbf{85.59}
      & \textbf{74.57}
      & \textbf{67.29}
      & \textbf{25.05} \\

    \cmidrule(lr){2-11}

      &
      \multirow{2}{*}{%
        \makecell[l]{Ours (9B,\\fine-tuned)}
      }
      & $\times$
      & 37.10
      & 15.11
      & 14.92
      & \textbf{1.1405}
      & 82.41
      & 70.93
      & 55.92
      & 30.00 \\

      &
      & $\checkmark$
      & \textbf{50.00}
      & \textbf{20.65}
      & \textbf{19.76}
      & 1.1571
      & \textbf{91.28}
      & \textbf{81.11}
      & \textbf{71.01}
      & \textbf{32.77} \\

    \bottomrule
  \end{tabularx}
\end{table*}

\begin{table*}[t]
  \centering
  \caption{Adapter composition for role-specialized training on
  Earth-Bench-Pro. All quality metrics are percentages, while Efficiency is a
  ratio.}
  \label{tab:exp-adapter-composition}

  \small
  \setlength{\tabcolsep}{3.0pt}
  \renewcommand{\arraystretch}{1.18}

  \begin{tabularx}{0.95\textwidth}{
    L{1.8cm}
    L{1.8cm}
    *{8}{Y}
  }
    \toprule
    \textbf{Planner}
      & \textbf{Executor}
      & \makecell[c]{\textbf{LLM-as-}\\\textbf{Judge}}
      & \mbox{\textbf{ROUGE-L}}
      & \textbf{EM}
      & \textbf{Efficiency}
      & \textbf{TAO}
      & \textbf{TIO}
      & \textbf{TEM}
      & \textbf{Param.} \\
    \midrule

    Base
      & Base
      & 38.31
      & 16.33
      & 15.73
      & \textbf{1.0129}
      & 85.59
      & 74.57
      & 67.29
      & 25.05 \\

    Base
      & Tuned
      & 43.15
      & 17.31
      & 18.15
      & \underline{0.9865}
      & 85.44
      & 72.24
      & 65.64
      & 28.34 \\

    Tuned
      & Base
      & \underline{44.76}
      & \textbf{20.67}
      & \underline{18.55}
      & 1.1839
      & \underline{89.47}
      & \underline{78.78}
      & \underline{70.09}
      & \underline{30.33} \\

    Tuned
      & Tuned
      & \textbf{50.00}
      & \underline{20.65}
      & \textbf{19.76}
      & 1.1571
      & \textbf{91.28}
      & \textbf{81.11}
      & \textbf{71.01}
      & \textbf{32.77} \\

    \bottomrule
  \end{tabularx}
\end{table*}

\textbf{Observation 1: domain skills improve reported quality metric.}
The improvement is particularly strong for trajectory fidelity. For the base
Earth-Agent-Pro model, adding skills raises TIO from 61.94\% to 74.57\% and TEM
from 51.71\% to 67.29\%. This pattern is consistent with skills constraining
the workflow search space toward expert processing procedures.

\textbf{Observation 2: skills and role-specialized training are
complementary.}
For the fine-tuned model, adding skills improves LLM-as-Judge accuracy by 12.90 points, ROUGE-L by 5.54 points, and EM by 4.84 points. It also raises TAO, TIO, TEM, and parameter accuracy by 8.87, 10.18, 15.09, and 2.77 points, respectively. These simultaneous gains across all metrics show that the external skill prior remains effective after role-specialized training. The two components operate at different levels: training strengthens the learned planning and execution policies, whereas skills provide explicit domain procedures that constrain workflow construction. Their combination consequently achieves the strongest answer-quality and trajectory-fidelity results, confirming that training does not make external skill guidance redundant.

\paragraph{Role-Specialized Training Ablation}

Finally, we disentangle the effects of Planner SFT and Executor GRPO through
adapter composition, planning-only evaluation, and oracle-plan execution
under carefully controlled component substitutions. Base and Tuned denote
the pretrained and role-specialized variants, respectively.

\textbf{Adapter composition.}
Table~\ref{tab:exp-adapter-composition} evaluates all four combinations of the
Planner and Executor adapters. Bold values indicate the best result and
underlined values indicate the runner-up; for Efficiency, values closer to
$1$ are better.

\begin{table}[!t]
  \centering

  \caption{Planning-only evaluation on Earth-Bench-Pro. Bold values indicate
  the better result within the Base/Tuned pair.}
  \label{tab:exp-planning-only}

  \small
  \setlength{\tabcolsep}{2.5pt}
  \renewcommand{\arraystretch}{1.18}

  \begin{tabularx}{0.95\columnwidth}{
    L{2.0cm}
    *{3}{Y}
  }
    \toprule
    \textbf{Planner}
      & \textbf{TAO}
      & \textbf{TIO}
      & \textbf{TEM} \\
    \midrule

    GPT-5
      & 88.86
      & 76.38
      & 70.03 \\

    \midrule

    InternLM3-8B
      & 72.15
      & 55.38
      & 46.87 \\
    GEMMA-4 (E4B-it)
      & 83.75
      & 71.34
      & 61.80 \\

    MiniMax-M2.7
      & 87.90
      & 71.89
      & 67.66 \\

    Base
      & 86.70
      & 71.87
      & 66.96 \\

    Tuned
      & \textbf{92.51}
      & \textbf{83.35}
      & \textbf{76.14} \\

    \bottomrule
  \end{tabularx}

  \par\vspace{2.0em}

  \caption{Oracle-plan Executor evaluation on Earth-Bench-Pro. Eff. denotes
  Efficiency. Bold values indicate the better result within the Base/Tuned
  pair.}
  \label{tab:exp-oracle-plan}

  \scriptsize
  \setlength{\tabcolsep}{1.2pt}
  \renewcommand{\arraystretch}{1.18}

  \begin{tabularx}{0.95\columnwidth}{
    L{1.75cm}
    *{5}{Y}
  }
    \toprule
    \textbf{Executor}
      & \makecell[c]{\textbf{LLM-as-}\\\textbf{Judge}}
      & \textbf{ROUGE-L}
      & \textbf{EM}
      & \textbf{Eff.}
      & \textbf{Parm.} \\
    \midrule

    GPT-5
      & 68.95
      & 24.31
      & 18.15
      & 0.9872
      & 39.58 \\

    \midrule

    InternLM3-8B
      & 38.71
      & 12.04
      & 10.48
      & 0.8620
      & 31.13 \\

    GEMMA-4 (E4B-it)
      & 33.87
      & 12.74
      & 10.89
      & 0.7109
      & 33.66 \\

    MiniMax-M2.7
      & 51.21
      & 19.44
      & 18.15
      & 0.8309
      & 35.72 \\

    Base
      & 46.37
      & 21.92
      & 20.16
      & 0.9081
      & 34.19 \\

    Tuned
      & \textbf{52.82}
      & \textbf{25.92}
      & \textbf{23.39}
      & \textbf{0.9274}
      & \textbf{37.20} \\

    \bottomrule
  \end{tabularx}
\end{table}

\textbf{Observation 1: Planner and Executor tuning provide complementary
benefits.}
Planner tuning produces the larger gains in workflow metrics, whereas Executor
tuning primarily improves final-answer quality and parameter grounding. Their
combination achieves the strongest overall Judge, EM, TAO, TIO, TEM, and
parameter scores.

\textbf{Observation 2: end-to-end parameter accuracy also depends on planning.}
The tuned-Planner/base-Executor configuration has higher parameter accuracy
than the base-Planner/tuned-Executor configuration. This occurs because the
end-to-end metric also depends on workflow alignment and execution progress,
motivating the isolated evaluations below.

\textbf{Planning-only isolates workflow composition.}

The planning-only evaluation isolates the Planner's prediction of workflow structure. We run only the four-role sequential debate and compare its output tool sequence directly with the reference workflow. This setting omits tool execution and Executor-side tool replacement, early termination, same-node retries, suffix replanning, and workflow editing. The resulting metrics measure static workflow composition because the planning-only trace retains proposed tools without execution, while the complete system records only verified tool calls. Planning-only scores can therefore exceed end-to-end trajectory scores when execution failures shorten the accepted trace.

\textbf{Observation 3: Planner SFT improves workflow composition.}
The tuned Planner reaches 92.51\% TAO, 83.35\% TIO, and 76.14\% TEM, improving over the base Planner by 5.81, 11.48, and 9.18 percentage points, respectively. It also exceeds GPT-5 by 3.65, 6.97, and 6.11 points on the same metrics. These gains isolate improved prediction of the reference tool set and order; end-to-end completion remains subject to subsequent execution.

\textbf{Oracle-plan execution isolates argument grounding.}

The oracle-plan evaluation isolates the Executor by fixing the workflow structure. We replace the predicted plan with the reference tool order and disable tool replacement, early termination, suffix replanning, and workflow editing. Each node still generates arguments, invokes the specified tool, verifies the returned observation, and retries the same tool when needed. This protocol fixes tool selection and ordering while preserving runtime argument grounding and environment interaction throughout the complete execution process.

\textbf{Observation 4: Executor GRPO improves argument grounding.}
Relative to the base Executor, the tuned Executor raises LLM-as-Judge accuracy from 46.37\% to 52.82\%, ROUGE-L from 21.92\% to 25.92\%, EM from 20.16\% to 23.39\%, and parameter accuracy from 34.19\% to 37.20\%. Efficiency also moves from 0.9081 to 0.9274, closer to the ideal value of $1$. GPT-5 still reaches 68.95\% LLM-as-Judge accuracy and 39.58\% parameter accuracy, indicating a remaining gap associated with backbone capability.

Together with the adapter-composition results in Table~\ref{tab:exp-adapter-composition}, these isolated evaluations show that Planner SFT improves workflow composition and Executor GRPO improves argument grounding under a fixed workflow. The combined system benefits from these distinct interface-specific gains.

\section{Discussion}
\label{sec:discussion}

\subsection{Qualitative Case Studies}
\label{sec:qualitative-cases}

\begin{figure*}[!t]
  \centering
  \includegraphics[width=0.95\textwidth]{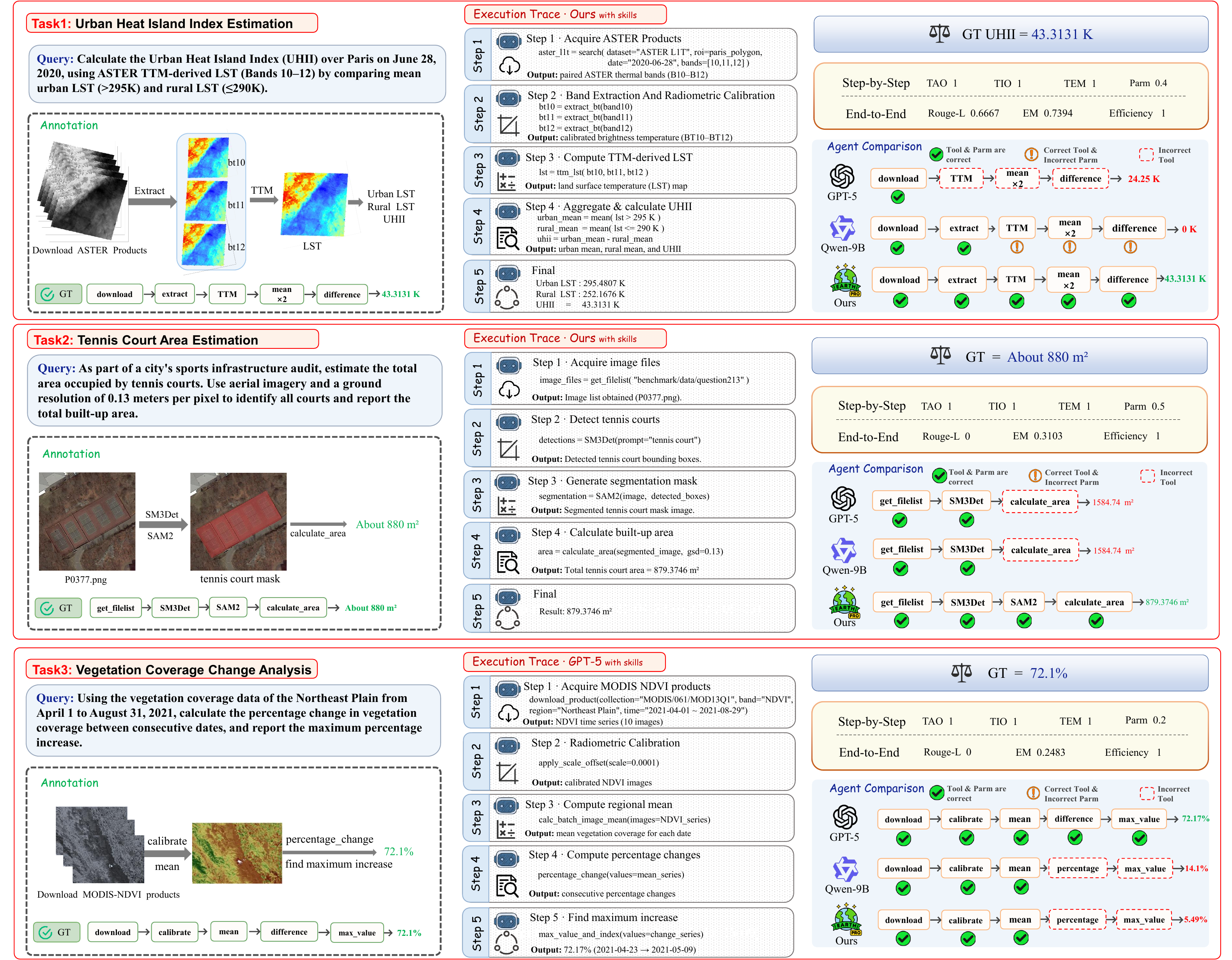}
\caption{\textbf{Representative execution cases on Earth-Bench-Pro.}
From top to bottom: Spectrum-based UHII estimation, RGB-based tennis-court area
estimation, and Product-based vegetation-coverage change analysis. The first
two cases show that our agent preserves the required tool dependencies and
matches the reference answers, whereas the third illustrates errors caused by
incorrect downstream tool selection.}
\label{fig:qualitative-cases}
\end{figure*}

We further analyze three representative execution traces covering spectral
observations, RGB imagery, and remote-sensing products, as shown in
Figure~\ref{fig:qualitative-cases}. These examples complement the aggregate
results by illustrating three factors that determine successful open-world EO
execution: workflow completeness, preservation of intermediate dependencies,
and correct grounding of tools and arguments.

In the Spectrum case, the agent must acquire ASTER products, extract
Bands 10--12, retrieve land-surface temperature using TTM, aggregate the urban
and rural temperatures, and calculate their difference. GPT-5 skips the band
extraction stage and applies the operations to incorrectly prepared
inputs, producing $24.25\,\mathrm{K}$. Qwen-9B includes extraction but
incorrectly grounds the thermal-processing and aggregation operations, resulting
in $0\,\mathrm{K}$. In contrast, our agent preserves the executable
dependency chain and recovers the reference UHII of
$43.3131\,\mathrm{K}$. This case shows that selecting plausible tools is
insufficient when their inputs and arguments are not grounded in the preceding
observations.

The RGB case further demonstrates the importance of intermediate
representations. GPT-5 and Qwen-9B detect the tennis courts but proceed directly
from the detections to area calculation. Without converting the coarse
detections into a pixel-level mask, both agents overestimate the occupied area
as $1584.74\,\mathrm{m}^{2}$. Our agent inserts SAM2 segmentation between
detection and measurement and calculates the area from the resulting mask using
the specified ground resolution. It therefore obtains
$879.3746\,\mathrm{m}^{2}$, consistent with the reference answer of
approximately $880\,\mathrm{m}^{2}$. The difference arises not from the final
measurement tool alone, but from whether the correct intermediate artifact is
constructed and passed downstream.

The Product case exposes a limitation. All three agents perform MODIS
acquisition, radiometric calibration, and regional-mean calculation. GPT-5
follows the reference change-analysis and maximum-value operations, obtaining
$72.17\%$, close to the reference value of $72.1\%$. Qwen-9B and our agent
instead select incorrect downstream operations, producing $14.1\%$ and
$5.49\%$, respectively. Thus, correct upstream data preparation does not
guarantee a correct answer when the scientific meaning of a downstream
operation is mismatched.

Taken together, these cases show that open-world EO execution requires more
than individually reasonable tool calls. Reliable quantitative answers depend
on maintaining the complete workflow, constructing the intended intermediate
representations, and grounding every downstream operation in both the accepted
execution evidence and the scientific objective. The Product failure also
indicates that skill guidance improves workflow reliability but does not fully
eliminate semantic tool-selection errors.

\subsection{Limitations and Trade-offs}

Open-world EO execution remains limited by exact workflow fidelity and numerical grounding. Even the strongest TEM and parameter scores in Table~\ref{tab:exp-v2-backbones} are 71.01\% and 32.77\%, respectively. The gap between reliable tool-set coverage and much lower parameter accuracy suggests that selecting relevant tools is easier than instantiating a fully faithful executable workflow. Open-ended QA removes answer candidates and requires the model or agent to directly produce a calibrated numerical value, which more closely reflects real EO requests for measurements. It also requires target localization, metric reasoning, unit handling, and correct answer formatting. The zero and near-zero Distance accuracy across most methods in Table~\ref{tab:exp-rsmllm} shows that precise open-ended spatial measurement remains an unresolved challenge for both remote-sensing MLLMs and current agent systems.

Workflow fidelity and final-task accuracy remain distinct, and capability profiles vary across models and modalities. In Table~\ref{tab:exp-openearth}, AFlow leads Operation F1 and ReAct leads image-generation accuracy. The fine-tuned 4B and 9B models achieve higher tool-order fidelity than Earth-Agent-Pro (GPT-5) but remain below it in answering accuracy, suggesting that final-answer synthesis remains more dependent on backbone capability. On Product tasks in Table~\ref{tab:exp-general-agents}, Codex reaches 75.00\% accuracy, compared with 55.00\% for the strongest Earth-Agent-Pro variant. This modality-specific result limits any claim of uniform superiority across all task families.

Accuracy improvements also incur additional execution cost. Table~\ref{tab:exp-general-agents} shows that fine-tuning the 9B variant raises average accuracy from 50.00\% to 58.33\% and latency from 67.83\,s to 111.79\,s. The fine-tuned small model remains faster than Earth-Agent-Pro (GPT-5), whose latency is 208.64\,s, while closing part of the accuracy gap. In Table~\ref{tab:exp-v2-frameworks}, Earth-Agent-Pro achieves an Efficiency ratio of 1.0911, whose distance to the ideal value of $1$ is smaller than that of ReAct, AFlow, or OpenEarthAgent. Its trajectories are slightly longer than the reference on average, whereas the baselines tend to terminate with fewer calls. Table~\ref{tab:exp-skill-ablation} further shows that adding skills moves Efficiency farther from $1$ in all three ablations, increasing the absolute deviation by 0.0596 for ReAct, 0.0084 for the base Earth-Agent-Pro model, and 0.0166 for the fine-tuned model. This modest length overhead accompanies consistent improvements in answer quality and trajectory fidelity and may reflect additional domain-recommended processing or verification steps, exposing a quality--length trade-off.

\subsection{Error Analysis}

\begin{figure}[t]
  \centering
  \includegraphics[width=\linewidth]{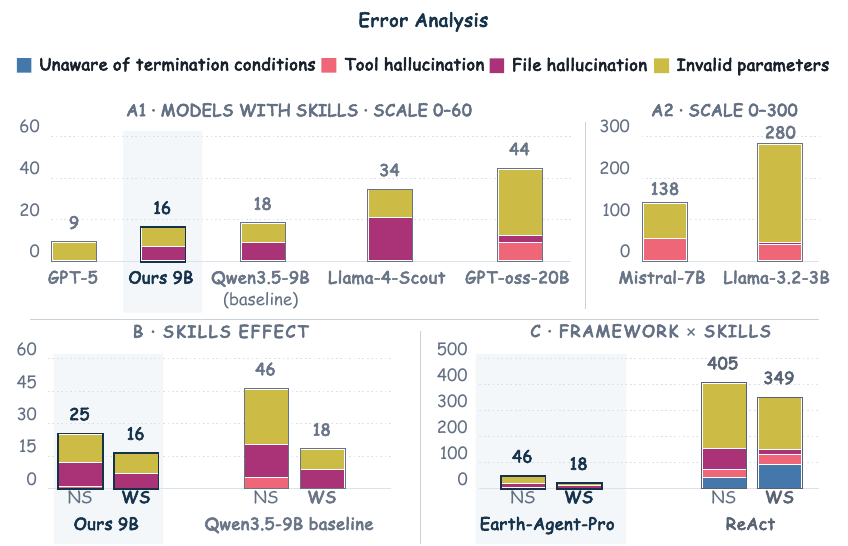}
  \caption{\textbf{Skills reduce operational error counts.}
  Each run contains 248 trajectories. Bars count logged operational errors, and one trajectory may contribute multiple events. Panel A compares with-skills (WS) counts across models, while Panel B compares no-skills (NS) with WS for the two 9B configurations. Panel C compares Earth-Agent-Pro and ReAct under the shared Qwen3.5-9B backbone. Colors denote repeated no-progress calls until the step limit, calls to nonexistent tools, accesses to nonexistent files or paths, and parameter schema, format, or value violations.}
  \label{fig:error_analysis}
\end{figure}

Figure~\ref{fig:error_analysis} shows how training, skill guidance, and framework design affect operational reliability. In Panel A1, role-specific training reduces file-hallucination events from nine to seven, while our fine-tuned Qwen3.5-9B model matches GPT-5 at nine logged invalid-parameter events. Panel A2 shows the backbone gap: Mistral-7B and Llama-3.2-3B retain 138 and 280 errors under WS, dominated by tool hallucinations and invalid parameters. Panel B shows that skills eliminate logged tool-hallucination events in both Qwen3.5-9B configurations and reduce total errors by 36.0\% for the fine-tuned model and 60.9\% for the base model. The larger base-model gain shows that skills transfer across configurations and remain effective without role-specific training. Panel C holds Qwen3.5-9B fixed and shows that Earth-Agent-Pro records fewer events than ReAct in every error category under NS and WS, lowering the total by 88.6\% and 94.8\%, respectively.


\section{Conclusion}

This work studies open-world EO execution, where an agent must locate runtime inputs, acquire observations when required, prepare data, perform domain computations, and generate an open-ended answer from runtime evidence given a high-level scientific question. Earth-Bench-Pro instantiates 248 expert-curated task cores as 744 matched questions, supporting comparisons across instruction following, autonomous planning, and open-world execution through executable evidence chains. Earth-Agent-Pro connects workflow planning with runtime operation grounding through expert skill guidance, workflow-centered structured memory, and localized suffix repair. Role-specialized training uses sequence-level SFT for Planner workflow composition and node-level GRPO for Executor argument grounding. On Earth-Bench-OW, Earth-Agent-Pro leads seven of eight framework-comparison metrics under a shared GPT-5 backbone and reaches 66.13\% LLM-as-Judge accuracy. Adapter composition and isolated evaluations record higher workflow-composition scores for the tuned Planner and higher answer and parameter scores for the tuned Executor under a fixed workflow.

The experiments also expose remaining constraints for open-world EO agents. Even the strongest configurations achieve higher tool coverage than exact workflow matching and parameter accuracy, while final-answer quality remains sensitive to backbone capability. Future work should improve argument grounding under runtime uncertainty, reduce the trajectory overhead introduced by skill-guided execution, and strengthen evidence synthesis in compact models.


\bibliographystyle{IEEEtran}
\bibliography{earthagentpp_refs}

@inproceedings{shu2026terrascope,
  title={TerraScope: Pixel-Grounded Visual Reasoning for Earth Observation},
  author={Shu, Yan and Ren, Bin and Xiong, Zhitong and Zhu, Xiao Xiang and Demir, Begum and Sebe, Nicu and Rota, Paolo},
  booktitle={Proceedings of the IEEE/CVF Conference on Computer Vision and Pattern Recognition},
  pages={16712--16722},
  year={2026}
}

@article{chen2025cangling,
  title={CangLing-KnowFlow: A Unified Knowledge-and-Flow-fused Agent for Comprehensive Remote Sensing Applications},
  author={Chen, Zhengchao and Wang, Haoran and Yao, Jing and Zhang, Jianshe and Ghamisi, Pedram and Zhou, Jun and Atkinson, Peter M and Zhang, Bing},
  journal={arXiv preprint arXiv:2512.15231},
  year={2025}
}

@article{shabbir2026openearthagent,
  title={OpenEarthAgent: A Unified Framework for Tool-Augmented Geospatial Agents},
  author={Shabbir, Akashah and Sheikh, Muhammad Umer and Munir, Muhammad Akhtar and Debary, Hiyam and Fiaz, Mustansar and Zaheer, Muhammad Zaigham and Fraccaro, Paolo and Khan, Fahad Shahbaz and Khan, Muhammad Haris and Zhu, Xiao Xiang and others},
  journal={arXiv preprint arXiv:2602.17665},
  year={2026}
}

@article{zhao2026openearth,
  title={OpenEarth-Agent: From Tool Calling to Tool Creation for Open-Environment Earth Observation},
  author={Zhao, Sijie and Liu, Feng and Zhang, Xueliang and Chen, Hao and Gu, Xinyu and Jiang, Zhe and Ling, Fenghua and Fei, Ben and Zhang, Wenlong and Wang, Junjue and others},
  journal={arXiv preprint arXiv:2603.22148},
  year={2026}
}

@article{ma2026eo,
  title={EO-Gym: A Multimodal, Interactive Environment for Earth Observation Agents},
  author={Ma, Sai and Li, Zhuang and Li, Sichao and Xu, Xinyue and Zhu, Ruibiao and Boston, Tony and Taylor, John A},
  journal={arXiv preprint arXiv:2605.01250},
  year={2026}
}

@article{nguyen2026terrabench,
  title={TerraBench: Can Agents Reason Over Heterogeneous Earth-System Data?},
  author={Nguyen, Dat Tien and Nguyen, Thao and Maani, Fadillah Adamsyah and Le, Huy M and Sheikh, Muhammad Umer and Saeed, Numan and Khan, Muhammad Haris and Khan, Salman},
  journal={arXiv preprint arXiv:2606.13148},
  year={2026}
}

@article{li2025designing,
  title={Designing Domain-Specific Agents via Hierarchical Task Abstraction Mechanism},
  author={Li, Kaiyu and Wang, Jiayu and Wang, Zhi and Qiao, Hui and Zhang, Weizhan and Meng, Deyu and Cao, Xiangyong},
  journal={arXiv preprint arXiv:2511.17198},
  year={2025}
}

@inproceedings{xiao2026geommbench,
  title={GeoMMBench and GeoMMAgent: Toward Expert-Level Multimodal Intelligence in Geoscience and Remote Sensing},
  author={Xiao, Aoran and Cheng, Shihao and Xu, Yonghao and Ren, Yexian and Chen, Hongruixuan and Yokoya, Naoto},
  booktitle={Proceedings of the IEEE/CVF Conference on Computer Vision and Pattern Recognition},
  pages={34843--34853},
  year={2026}
}

@article{zhan2025skyeyegpt,
  title={SkyEyeGPT: Unifying Remote Sensing Vision-Language Tasks via Instruction Tuning with Large Language Model},
  author={Zhan, Yang and Xiong, Zhitong and Yuan, Yuan},
  journal={ISPRS Journal of Photogrammetry and Remote Sensing},
  volume={221},
  pages={64--77},
  year={2025},
  publisher={Elsevier}
}

@article{hong2024spectralgpt,
  title={SpectralGPT: Spectral Remote Sensing Foundation Model},
  author={Hong, Danfeng and Zhang, Bing and Li, Xuyang and Li, Yuxuan and Li, Chenyu and Yao, Jing and Yokoya, Naoto and Li, Hao and Ghamisi, Pedram and Jia, Xiuping and others},
  journal={IEEE Transactions on Pattern Analysis and Machine Intelligence},
  volume={46},
  number={8},
  pages={5227--5244},
  year={2024},
  publisher={IEEE}
}

@article{diao2025ringmo,
  title={RingMo-Aerial: An Aerial Remote Sensing Foundation Model with Affine Transformation Contrastive Learning},
  author={Diao, Wenhui and Yu, Haichen and Kang, Kaiyue and Ling, Tong and Liu, Di and Feng, Yingchao and Bi, Hanbo and Ren, Libo and Li, Xuexue and Mao, Yongqiang and others},
  journal={IEEE Transactions on Pattern Analysis and Machine Intelligence},
  year={2025},
  publisher={IEEE}
}

@article{chen2025rest,
  title={REST: Holistic Learning for End-to-End Semantic Segmentation of Whole-Scene Remote Sensing Imagery},
  author={Chen, Wei and Bruzzone, Lorenzo and Dang, Bo and Gao, Yuan and Deng, Youming and Yu, Jin-Gang and Yuan, Liangqi and Li, Yansheng},
  journal={IEEE Transactions on Pattern Analysis and Machine Intelligence},
  year={2025},
  publisher={IEEE}
}

@article{bi2025ringmoe,
  title={RingMoE: Mixture-of-Modality-Experts Multi-Modal Foundation Models for Universal Remote Sensing Image Interpretation},
  author={Bi, Hanbo and Feng, Yingchao and Tong, Boyuan and Wang, Mengyu and Yu, Haichen and Mao, Yongqiang and Chang, Hao and Diao, Wenhui and Wang, Peijin and Yu, Yue and others},
  journal={IEEE Transactions on Pattern Analysis and Machine Intelligence},
  year={2025},
  publisher={IEEE}
}

@article{gong2025crossearth,
  title={CrossEarth: Geospatial Vision Foundation Model for Domain Generalizable Remote Sensing Semantic Segmentation},
  author={Gong, Ziyang and Wei, Zhixiang and Wang, Di and Hu, Xiaoxing and Ma, Xianzheng and Chen, Hongruixuan and Jia, Yuru and Deng, Yupeng and Ji, Zhenming and Zhu, Xiangwei and others},
  journal={IEEE Transactions on Pattern Analysis and Machine Intelligence},
  year={2025},
  publisher={IEEE}
}

@article{wang2025hypersigma,
  title={HyperSIGMA: Hyperspectral Intelligence Comprehension Foundation Model},
  author={Wang, Di and Hu, Meiqi and Jin, Yao and Miao, Yuchun and Yang, Jiaqi and Xu, Yichu and Qin, Xiaolei and Ma, Jiaqi and Sun, Lingyu and Li, Chenxing and others},
  journal={IEEE Transactions on Pattern Analysis and Machine Intelligence},
  year={2025},
  publisher={IEEE}
}

@inproceedings{kuckreja2024geochat,
  title={GeoChat: Grounded Large Vision-Language Model for Remote Sensing},
  author={Kuckreja, Kartik and Danish, Muhammad Sohail and Naseer, Muzammal and Das, Abhijit and Khan, Salman and Khan, Fahad Shahbaz},
  booktitle={Proceedings of the IEEE/CVF Conference on Computer Vision and Pattern Recognition},
  pages={27831--27840},
  year={2024}
}

@inproceedings{muhtar2024lhrs,
  title={LHRS-Bot: Empowering Remote Sensing with VGI-Enhanced Large Multimodal Language Model},
  author={Muhtar, Dilxat and Li, Zhenshi and Gu, Feng and Zhang, Xueliang and Xiao, Pengfeng},
  booktitle={European Conference on Computer Vision},
  pages={440--457},
  year={2024},
  organization={Springer}
}

@inproceedings{pang2025vhm,
  title={VHM: Versatile and Honest Vision Language Model for Remote Sensing Image Analysis},
  author={Pang, Chao and Weng, Xingxing and Wu, Jiang and Li, Jiayu and Liu, Yi and Sun, Jiaxing and Li, Weijia and Wang, Shuai and Feng, Litong and Xia, Gui-Song and others},
  booktitle={Proceedings of the AAAI Conference on Artificial Intelligence},
  volume={39},
  number={6},
  pages={6381--6388},
  year={2025}
}

@article{lobry2020rsvqa,
  title={RSVQA: Visual Question Answering for Remote Sensing Data},
  author={Lobry, Sylvain and Marcos, Diego and Murray, Jesse and Tuia, Devis},
  journal={IEEE Transactions on Geoscience and Remote Sensing},
  volume={58},
  number={12},
  pages={8555--8566},
  year={2020},
  publisher={IEEE}
}

@article{li2024vrsbench,
  title={Vrsbench: A versatile vision-language benchmark dataset for remote sensing image understanding},
  author={Li, Xiang and Ding, Jian and Elhoseiny, Mohamed},
  journal={Advances in Neural Information Processing Systems},
  volume={37},
  pages={3229--3242},
  year={2024}
}

@inproceedings{danish2025geobench,
  title={GeoBench-VLM: Benchmarking Vision-Language Models for Geospatial Tasks},
  author={Danish, Muhammad and Munir, Muhammad Akhtar and Shah, Syed Roshaan Ali and Kuckreja, Kartik and Khan, Fahad Shahbaz and Fraccaro, Paolo and Lacoste, Alexandre and Khan, Salman},
  booktitle={Proceedings of the IEEE/CVF International Conference on Computer Vision},
  pages={7132--7142},
  year={2025}
}

@inproceedings{wang2025xlrs,
  title={XLRS-Bench: Could Your Multimodal LLMs Understand Extremely Large Ultra-High-Resolution Remote Sensing Imagery?},
  author={Wang, Fengxiang and Wang, Hongzhen and Guo, Zonghao and Wang, Di and Wang, Yulin and Chen, Mingshuo and Ma, Qiang and Lan, Long and Yang, Wenjing and Zhang, Jing and others},
  booktitle={Proceedings of the IEEE/CVF Conference on Computer Vision and Pattern Recognition},
  pages={14325--14336},
  year={2025}
}

@article{wang2026disasterm3,
  title={Disasterm3: A remote sensing vision-language dataset for disaster damage assessment and response},
  author={Wang, Junjue and Xuan, Weihao and Qi, Heli and Liu, Zhihao and Liu, Kunyi and Wu, Yuhan and Chen, Hongruixuan and Song, Jian and Xia, Junshi and Zheng, Zhuo and others},
  journal={Advances in Neural Information Processing Systems},
  volume={38},
  year={2026}
}

@article{shabbir2025thinkgeo,
  title={Thinkgeo: Evaluating tool-augmented agents for remote sensing tasks},
  author={Shabbir, Akashah and Munir, Muhammad Akhtar and Dudhane, Akshay and Sheikh, Muhammad Umer and Khan, Muhammad Haris and Fraccaro, Paolo and Moreno, Juan Bernabe and Khan, Fahad Shahbaz and Khan, Salman},
  journal={arXiv preprint arXiv:2505.23752},
  year={2025}
}

@inproceedings{feng2025earth,
  title     = {{Earth-Agent}: Unlocking the Full Landscape of Earth Observation with Agents},
  author    = {Feng, Peilin and Lv, Zhutao and Ye, Junyan and Wang, Xiaolei and
               Huo, Xinjie and Yu, Jinhua and Xu, Wanghan and Zhang, Wenlong and
               Bai, Lei and He, Conghui and Li, Weijia},
  booktitle = {International Conference on Learning Representations},
  year      = {2026},
  url       = {https://openreview.net/forum?id=dkIXAbWuxO}
}

@article{xu2024rs,
  title   = {{RS-Agent}: Automating Remote Sensing Tasks through Intelligent Agent},
  author  = {Xu, Wenjia and Yu, Zijian and Mu, Boyang and Wang, Jiuniu and
             Wei, Zhiwei and Peng, Mugen},
  journal = {Science China Information Sciences},
  volume  = {69},
  number  = {8},
  pages   = {180302},
  year    = {2026},
  doi     = {10.1007/s11432-026-5026-5},
  url     = {https://doi.org/10.1007/s11432-026-5026-5}
}

@article{liu2024change,
  title={Change-agent: Toward interactive comprehensive remote sensing change interpretation and analysis},
  author={Liu, Chenyang and Chen, Keyan and Zhang, Haotian and Qi, Zipeng and Zou, Zhengxia and Shi, Zhenwei},
  journal={IEEE Transactions on Geoscience and Remote Sensing},
  volume={62},
  pages={1--16},
  year={2024},
  publisher={IEEE}
}

@inproceedings{kao2026towards,
  title={Towards llm agents for earth observation},
  author={Kao, Chia Hsiang and Zhao, Wenting and Lam, Cheryl and Umap, Aarush and Revankar, Shreelekha and Speas, Samuel and Bhagat, Snehal and Datta, Rajeev and Phoo, Cheng Perng and Mall, Utkarsh and others},
  booktitle={Findings of the Association for Computational Linguistics: ACL 2026},
  pages={2597--2611},
  year={2026}
}

@inproceedings{bhattaram2025geoflow,
  title={GeoFlow: Agentic Workflow Automation for Geospatial Tasks},
  author={Bhattaram, Amulya and Chung, Justin and Chung, Stanley and Gupta, Ranit and Ramamoorthy, Janani and Gullapalli, Kartikeya and Marculescu, Diana and Stamoulis, Dimitrios},
  booktitle={Proceedings of the 33rd ACM International Conference on Advances in Geographic Information Systems},
  pages={1150--1153},
  year={2025}
}

@article{dai2026geoevolver,
  title={Experience-Driven Multi-Agent Systems Are Training-Free Context-Aware Earth Observers},
  author={Dai, Pengyu and Xuan, Weihao and Wang, Junjue and Chen, Hongruixuan and Song, Jian and Ou, Yafei and Yokoya, Naoto},
  journal={arXiv preprint arXiv:2602.02559},
  year={2026}
}

@article{remoteagent2026,
  title={RemoteAgent: Bridging Vague Human Intents and Earth Observation with RL-Based Agentic MLLMs},
  author={Yao, Liang and Xu, Shengxiang and Liu, Fan and Zhang, Chuanyi and Yao, Bishun and Min, Rui and Li, Yongjun and Ouyang, Chaoqian and Di, Shimin and Zhang, Min-Ling},
  journal={arXiv preprint arXiv:2604.07765},
  year={2026}
}

@inproceedings{yao2022react,
  title     = {{ReAct}: Synergizing Reasoning and Acting in Language Models},
  author    = {Yao, Shunyu and Zhao, Jeffrey and Yu, Dian and Du, Nan and
               Shafran, Izhak and Narasimhan, Karthik and Cao, Yuan},
  booktitle = {International Conference on Learning Representations},
  year      = {2023},
  url       = {https://openreview.net/forum?id=WE_vluYUL-X}
}

@article{patil2024gorilla,
  title={Gorilla: Large language model connected with massive apis},
  author={Patil, Shishir G and Zhang, Tianjun and Wang, Xin and Gonzalez, Joseph E},
  journal={Advances in Neural Information Processing Systems},
  volume={37},
  pages={126544--126565},
  year={2024}
}

@inproceedings{qin2024toolllm,
  title={Toolllm: Facilitating large language models to master 16000+ real-world apis},
  author={Qin, Yujia and Liang, Shihao and Ye, Yining and Zhu, Kunlun and Yan, Lan and Lu, Yaxi and Lin, Yankai and Cong, Xin and Tang, Xiangru and Qian, Bill and others},
  booktitle={International Conference on Learning Representations},
  volume={2024},
  pages={9695--9717},
  year={2024}
}

@article{wang2023voyager,
  title   = {{Voyager}: An Open-Ended Embodied Agent with Large Language Models},
  author  = {Wang, Guanzhi and Xie, Yuqi and Jiang, Yunfan and
             Mandlekar, Ajay and Xiao, Chaowei and Zhu, Yuke and
             Fan, Linxi and Anandkumar, Anima},
  journal = {Transactions on Machine Learning Research},
  year    = {2024},
  url     = {https://openreview.net/forum?id=ehfRiF0R3a}
}

@article{hu2025rsgpt,
  title={RSGPT: A Remote Sensing Vision Language Model and Benchmark},
  author={Hu, Yuan and Yuan, Jianlong and Wen, Congcong and Lu, Xiaonan and Liu, Yu and Li, Xiang},
  journal={ISPRS Journal of Photogrammetry and Remote Sensing},
  volume={224},
  pages={272--286},
  year={2025},
  publisher={Elsevier}
}

@article{luo2024skysensegpt,
  title={SkySenseGPT: A Fine-Grained Instruction Tuning Dataset and Model for Remote Sensing Vision-Language Understanding},
  author={Luo, Junwei and Pang, Zhen and Zhang, Yongjun and Wang, Tingzhu and Wang, Linlin and Dang, Bo and Lao, Jiangwei and Wang, Jian and Chen, Jingdong and Tan, Yihua and others},
  journal={arXiv preprint arXiv:2406.10100},
  year={2024}
}

@article{zhang2024earthgpt,
  title={EarthGPT: A universal multimodal large language model for multisensor image comprehension in remote sensing domain},
  author={Zhang, Wei and Cai, Miaoxin and Zhang, Tong and Zhuang, Yin and Mao, Xuerui},
  journal={IEEE Transactions on Geoscience and Remote Sensing},
  volume={62},
  pages={1--20},
  year={2024},
  publisher={IEEE}
}

@article{zhang2024earthmarker,
  title={EarthMarker: A visual prompting multimodal large language model for remote sensing},
  author={Zhang, Wei and Cai, Miaoxin and Zhang, Tong and Zhuang, Yin and Li, Jun and Mao, Xuerui},
  journal={IEEE Transactions on Geoscience and Remote Sensing},
  volume={63},
  pages={1--19},
  year={2024},
  publisher={IEEE}
}

@article{zhang2025earthgpt,
  title   = {{EarthGPT-X}: A Spatial {MLLM} for Multilevel Multisource Remote
             Sensing Imagery Understanding With Visual Prompting},
  author  = {Zhang, Wei and Cai, Miaoxin and Ning, Yaqian and Zhang, Tong and
             Zhuang, Yin and Lu, Shijian and Chen, He and Li, Jun and Mao, Xuerui},
  journal = {IEEE Transactions on Geoscience and Remote Sensing},
  volume  = {63},
  pages   = {1--21},
  year    = {2025},
  doi     = {10.1109/TGRS.2025.3626941},
  url     = {https://doi.org/10.1109/TGRS.2025.3626941}
}

@inproceedings{soni2025earthdial,
  title={Earthdial: Turning multi-sensory earth observations to interactive dialogues},
  author={Soni, Sagar and Dudhane, Akshay and Debary, Hiyam and Fiaz, Mustansar and Munir, Muhammad Akhtar and Danish, Muhammad Sohail and Fraccaro, Paolo and Watson, Campbell D and Klein, Levente J and Khan, Fahad Shahbaz and others},
  booktitle={Proceedings of the Computer Vision and Pattern Recognition Conference},
  pages={14303--14313},
  year={2025}
}

@article{shabbir2025geopixel,
  title={GeoPixel: Pixel Grounding Large Multimodal Model in Remote Sensing},
  author={Shabbir, Akashah and Zumri, Mohammed and Bennamoun, Mohammed and Khan, Fahad S. and Khan, Salman},
  journal={Proceedings of the International Conference on Machine Learning},
  year={2025}
}

@article{an2026choice,
  title={Choice: benchmarking the remote sensing capabilities of large vision-language models},
  author={An, Xiao and Sun, Jiaxing and Gui, Zihan and He, Wei},
  journal={Advances in Neural Information Processing Systems},
  volume={38},
  year={2026}
}

@inproceedings{zhang2024good,
  title={Good at captioning bad at counting: Benchmarking gpt-4v on earth observation data},
  author={Zhang, Chenhui and Wang, Sherrie},
  booktitle={Proceedings of the IEEE/CVF Conference on Computer Vision and Pattern Recognition},
  pages={7839--7849},
  year={2024}
}

@inproceedings{zhou2025urbench,
  title={URBench: A Comprehensive Benchmark for Evaluating Large Multimodal Models in Multi-View Urban Scenarios},
  author={Zhou, Baichuan and Yang, Haote and Chen, Dairong and Ye, Junyan and Bai, Tianyi and Yu, Jinhua and Zhang, Songyang and Lin, Dahua and He, Conghui and Li, Weijia},
  booktitle={Proceedings of the AAAI Conference on Artificial Intelligence},
  volume={39},
  number={10},
  pages={10707--10715},
  year={2025}
}

@inproceedings{guo2024rschatgpt,
  title={Remote Sensing ChatGPT: Solving Remote Sensing Tasks with ChatGPT and Visual Models},
  author={Guo, Haonan and Su, Xin and Wu, Chen and Du, Bo and Zhang, Liangpei and Li, Deren},
  booktitle={IGARSS 2024--2024 IEEE International Geoscience and Remote Sensing Symposium},
  pages={11474--11478},
  year={2024},
  organization={IEEE}
}

@article{hu2025ringmoagent,
  title   = {{RingMo-Agent}: A Unified Remote Sensing Foundation Model
             for Multi-Platform and Multi-Modal Reasoning},
  author  = {Hu, Huiyang and Wang, Peijin and Feng, Yingchao and Wei, Kaiwen and
             Yin, Wenxin and Diao, Wenhui and Wang, Mengyu and Bi, Hanbo and
             Kang, Kaiyue and Ling, Tong and Fu, Kun and Sun, Xian},
  journal = {IEEE Transactions on Pattern Analysis and Machine Intelligence},
  year    = {2026},
  note    = {Early Access},
  doi     = {10.1109/TPAMI.2026.3718699},
  url     = {https://doi.org/10.1109/TPAMI.2026.3718699}
}

@article{guo2025earthlink,
  title={EarthLink: A Self-Evolving AI Agent for Climate Science},
  author={Guo, Zijie and Wang, Jiong and Yue, Xiaoyu and Wei, Wangxu and Jiang, Zhe and Xu, Wanghan and Fei, Ben and Zhang, Wenlong and Gu, Xinyu and Cheng, Lijing and others},
  journal={arXiv preprint arXiv:2507.17311},
  year={2025}
}

@inproceedings{irvin2025teochat,
  title={Teochat: A large vision-language assistant for temporal earth observation data},
  author={Irvin, Jeremy and Liu, Emily and Chen, Joyce and Dormoy, Ines and Kim, Jinyoung and Khanna, Samar and Zheng, Zhuo and Ermon, Stefano},
  booktitle={International Conference on Learning Representations},
  volume={2025},
  pages={68883--68911},
  year={2025}
}

@article{shu2025earthmind,
  title={Earthmind: Towards multi-granular and multi-sensor earth observation with large multimodal models},
  author={Shu, Yan and Ren, Bin and Xiong, Zhitong and Pani Paudel, Danda and Van Gool, Luc and Demir, Begum and Sebe, Nicu and Rota, Paolo},
  journal={arXiv e-prints},
  pages={arXiv--2506},
  year={2025}
}

@misc{shao2024deepseekmath,
      title={DeepSeekMath: Pushing the Limits of Mathematical Reasoning in Open Language Models}, 
      author={Zhihong Shao and Peiyi Wang and Qihao Zhu and Runxin Xu and Junxiao Song and Xiao Bi and Haowei Zhang and Mingchuan Zhang and Y. K. Li and Y. Wu and Daya Guo},
      year={2024},
      eprint={2402.03300},
      archivePrefix={arXiv},
      primaryClass={cs.CL},
      url={https://arxiv.org/abs/2402.03300}, 
}

@article{yu2026geoagentbench,
  title={GeoAgentBench: A Dynamic Execution Benchmark for Tool-Augmented Agents in Spatial Analysis},
  author={Yu, Bo and Yang, Cheng and Hou, Dongyang and Liu, Chengfu and Liu, Jiayao and Wang, Chi and Zhang, Zhiming and Li, Haifeng and Yang, Wentao},
  journal={arXiv preprint arXiv:2604.13888},
  year={2026}
}

@article{wang2026dora,
  title={{Can LLM Agents Respond to Disasters? Benchmarking Heterogeneous Geospatial Reasoning in Emergency Operations}},
  author={Wang, Junjue and Xuan, Weihao and Qi, Heli and Dai, Pengyu and Liu, Kunyi and Chen, Hongruixuan and Zheng, Zhuo and Xia, Junshi and Ermon, Stefano and Yokoya, Naoto},
  journal={arXiv preprint arXiv:2605.11633},
  year={2026}
}

@article{yan2026terralogic,
  title={TerraLogic: A Benchmark for Hierarchical Geospatial Reasoning in Earth Observation},
  author={Yan, Yuhang and Mou, Linchao and Yang, Bokang and Li, Qingyu},
  journal={arXiv preprint arXiv:2607.12497},
  year={2026}
}

@article{yu2026nasaeobench,
  title={Bringing Agentic Search to Earth Observation Data Discovery},
  author={Yu, Minghan and Sun, Youran and Yi, Chugang and Wen, Yixin and Yang, Haizhao},
  journal={arXiv preprint arXiv:2607.02387},
  year={2026}
}

@inproceedings{krechetova2025geobenchx,
  title={GeoBenchX: Benchmarking LLMs in agent solving multistep geospatial tasks},
  author={Krechetova, Varvara and Kochedykov, Denis},
  booktitle={Proceedings of the 1st ACM SIGSPATIAL International Workshop on Generative and Agentic AI for Multi-Modality Space-Time Intelligence},
  pages={27--35},
  year={2025}
}

@article{liu2025rescueadi,
  title={RescueADI: Adaptive disaster interpretation in remote sensing images with autonomous agents},
  author={Liu, Zhuoran and Zhao, Danpei and Yuan, Bo and Jiang, Zhiguo},
  journal={IEEE Transactions on Geoscience and Remote Sensing},
  volume={63},
  pages={1--14},
  year={2025},
  publisher={IEEE}
}

@inproceedings{singh2024geollm,
  title={Geollm-engine: A realistic environment for building geospatial copilots},
  author={Singh, Simranjit and Fore, Michael and Stamoulis, Dimitrios},
  booktitle={2024 IEEE/CVF Conference on Computer Vision and Pattern Recognition Workshops (CVPRW)},
  pages={585--594},
  year={2024},
  organization={IEEE}
}

@article{wang2026towards,
  title={Towards realistic earth-observation constellation scheduling: Benchmark and methodology},
  author={Wang, Luting and Xiang, Yinghao and Huang, Hongliang and Li, Dongjun and Gao, Chen and Liu, Si},
  journal={Advances in Neural Information Processing Systems},
  volume={38},
  pages={85923--85944},
  year={2026}
}

@inproceedings{zhang2025aflow,
  title={Aflow: Automating agentic workflow generation},
  author={Zhang, Jiayi and Xiang, Jinyu and Yu, Zhaoyang and Teng, Fengwei and Chen, Xionghui and Chen, Jiaqi and Zhuge, Mingchen and Cheng, Xin and Hong, Sirui and Wang, Jinlin and others},
  booktitle={International Conference on Learning Representations},
  volume={2025},
  pages={34040--34077},
  year={2025}
}

@inproceedings{hu2021lora,
  title     = {{LoRA}: Low-Rank Adaptation of Large Language Models},
  author    = {Hu, Edward J. and Shen, Yelong and Wallis, Phillip and
               Allen-Zhu, Zeyuan and Li, Yuanzhi and Wang, Shean and
               Wang, Lu and Chen, Weizhu},
  booktitle = {International Conference on Learning Representations},
  year      = {2022},
  url       = {https://openreview.net/forum?id=nZeVKeeFYf9}
}

@article{ye2025satellite,
  title={Satellite image synthesis from street view with fine-grained spatial textual guidance: A novel framework},
  author={Ye, Junyan and He, Jun and Zhang, Xiang and Lin, Yi and Lin, Honglin and He, Conghui and Li, Weijia},
  journal={IEEE Geoscience and Remote Sensing Magazine},
  volume={13},
  number={3},
  pages={395--414},
  year={2025},
  publisher={IEEE}
}

@inproceedings{ye2024skydiffusion,
  title     = {Leveraging {BEV} Paradigm for Ground-to-Aerial Image Synthesis},
  author    = {Ye, Junyan and He, Jun and Li, Weijia and Lv, Zhutao and
               Lin, Yi and Yu, Jinhua and Yang, Haote and He, Conghui},
  booktitle = {Proceedings of the IEEE/CVF International Conference on Computer Vision},
  pages     = {28451--28461},
  year      = {2025},
  url       = {https://openaccess.thecvf.com/content/ICCV2025/html/Ye_Leveraging_BEV_Paradigm_for_Ground-to-Aerial_Image_Synthesis_ICCV_2025_paper.html}
}

@inproceedings{ye2024sg,
  title={SG-BEV: Satellite-guided BEV fusion for cross-view semantic segmentation},
  author={Ye, Junyan and Luo, Qiyan and Yu, Jinhua and Zhong, Huaping and Zheng, Zhimeng and He, Conghui and Li, Weijia},
  booktitle={2024 IEEE/CVF Conference on Computer Vision and Pattern Recognition (CVPR)},
  pages={27748--27757},
  year={2024},
  organization={IEEE}
}

@article{li2024crossviewdiff,
  title={Crossviewdiff: A cross-view diffusion model for satellite-to-street view synthesis},
  author={Li, Weijia and He, Jun and Ye, Junyan and Zhong, Huaping and Zheng, Zhimeng and Huang, Zilong and Lin, Dahua and He, Conghui},
  journal={arXiv preprint arXiv:2408.14765},
  year={2024}
}

@article{anderson2017earth,
  title={Earth observation in service of the 2030 Agenda for Sustainable Development},
  author={Anderson, Katherine and Ryan, Barbara and Sonntag, William and Kavvada, Argyro and Friedl, Lawrence},
  journal={Geo-spatial Information Science},
  volume={20},
  number={2},
  pages={77--96},
  year={2017},
  publisher={Taylor \& Francis}
}

@article{li2024roadcorrector,
  title={RoadCorrector: A structure-aware road extraction method for road connectivity and topology correction},
  author={Li, Jinpeng and He, Jun and Li, Weijia and Chen, Jiabin and Yu, Jinhua},
  journal={IEEE Transactions on Geoscience and Remote Sensing},
  volume={62},
  pages={1--18},
  year={2024},
  publisher={IEEE}
}

@article{brown2025alphaearth,
  title={Alphaearth foundations: An embedding field model for accurate and efficient global mapping from sparse label data},
  author={Brown, Christopher F and Kazmierski, Michal R and Pasquarella, Valerie J and Rucklidge, William J and Samsikova, Masha and Zhang, Chenhui and Shelhamer, Evan and Lahera, Estefania and Wiles, Olivia and Ilyushchenko, Simon and others},
  journal={arXiv preprint arXiv:2507.22291},
  year={2025}
}

@article{liu2020large,
  title={Large-scale crop mapping from multisource remote sensing images in google earth engine},
  author={Liu, Xinkai and Zhai, Han and Shen, Yonglin and Lou, Benke and Jiang, Changmin and Li, Tianqi and Hussain, Sayed Bilal and Shen, Guoling},
  journal={IEEE Journal of Selected Topics in Applied Earth Observations and Remote Sensing},
  volume={13},
  pages={414--427},
  year={2020},
  publisher={IEEE}
}

@article{transon2018survey,
  title={Survey of hyperspectral earth observation applications from space in the sentinel-2 context},
  author={Transon, Julie and d’Andrimont, Rapha{\"e}l and Maugnard, Alexandre and Defourny, Pierre},
  journal={Remote Sensing},
  volume={10},
  number={2},
  pages={157},
  year={2018},
  publisher={MDPI}
}

@article{kokkoris2024role,
  title={The role of Earth observation in ecosystem accounting: A review of advances, challenges and future directions},
  author={Kokkoris, Ioannis P and Smets, Bruno and Hein, Lars and Mallinis, Giorgos and Buchhorn, Marcel and Balbi, Stefano and {\v{C}}erneck{\`y}, J{\'a}n and Paganini, Marc and Dimopoulos, Panayotis},
  journal={Ecosystem Services},
  volume={70},
  pages={101659},
  year={2024},
  publisher={Elsevier}
}

@article{Sudmanns02012018,
author = {Martin Sudmanns and Dirk Tiede and Stefan Lang and Andrea Baraldi},
title = {Semantic and syntactic interoperability in online processing of big Earth observation data},
journal = {International Journal of Digital Earth},
volume = {11},
number = {1},
pages = {95--112},
year = {2018},
publisher = {Taylor \& Francis},
doi = {10.1080/17538947.2017.1332112},

    note ={PMID: 29387171},


URL = { 
    
        https://doi.org/10.1080/17538947.2017.1332112
    
    

},
eprint = { 
    
        https://doi.org/10.1080/17538947.2017.1332112
    
    

}

}

@article{WEISE2020111892,
title = {Wetland extent tools for SDG 6.6.1 reporting from the Satellite-based Wetland Observation Service (SWOS)},
journal = {Remote Sensing of Environment},
volume = {247},
pages = {111892},
year = {2020},
issn = {0034-4257},
doi = {https://doi.org/10.1016/j.rse.2020.111892},
url = {https://www.sciencedirect.com/science/article/pii/S0034425720302625},
author = {Kathrin Weise and Rene Höfer and Jonas Franke and Anis Guelmami and Will Simonson and Javier Muro and Brian O’Connor and Adrian Strauch and Stephan Flink and Jonas Eberle and Eric Mino and Susanne Thulin and Petra Philipson and Eric {van Valkengoed} and John Truckenbrodt and Franziska Zander and Antonio Sánchez and Christoph Schröder and Frank Thonfeld and Eleni Fitoka and Emma Scott and Matthew Ling and Michael Schwarz and Ina Kunz and Grit Thürmer and Anouska Plasmeijer and Lammert Hilarides}
}

@article{FENG2025104825,
title = {Towards a barrier-free GeoQA portal: Natural language interaction with geospatial data using multi-agent LLMs and semantic search},
journal = {International Journal of Applied Earth Observation and Geoinformation},
volume = {144},
pages = {104825},
year = {2025},
issn = {1569-8432},
doi = {https://doi.org/10.1016/j.jag.2025.104825},
url = {https://www.sciencedirect.com/science/article/pii/S1569843225004728},
author = {Yu Feng and Puzhen Zhang and Guohui Xiao and Linfang Ding and Liqiu Meng}
}

@inproceedings{lin2004rouge,
  title={Rouge: A package for automatic evaluation of summaries},
  author={Lin, Chin-Yew},
  booktitle={Text summarization branches out},
  pages={74--81},
  year={2004}
}

@article{wang2025omniearth,
  title={OmniEarth-Bench: Towards Holistic Evaluation of Earth's Six Spheres and Cross-Spheres Interactions with Multimodal Observational Earth Data},
  author={Wang, Fengxiang and Chen, Mingshuo and He, Xuming and Zhang, Yi-Fan and Li, Yueying and Liu, Feng and Guo, Zijie and Hu, Zhenghao and Wang, Jiong and Xu, Jingyi and others},
  journal={arXiv preprint arXiv:2505.23522},
  year={2025}
}

@article{li2026can,
  title={Can large multimodal models understand agricultural scenes? benchmarking with agromind},
  author={Li, Qingmei and Zhang, Yang and Mai, Zurong and Chen, Yuhang and Huang, Henglian and Zhang, Jiarui and Zhang, Zhiwei and Wen, Yibin and Li, Weijia and Fu, Haohuan and others},
  journal={Advances in Neural Information Processing Systems},
  volume={38},
  year={2026}
}

@article{zhang2026a2mae,
  title={A 2-MAE: A spatial-temporal-spectral unified remote sensing pre-training method based on anchor-aware masked autoencoder},
  author={Zhang, Lixian and Zhao, Yi and Dong, Runmin and Zhang, Jinxiao and Yuan, Shuai and Cao, Shilei and Chen, Mengxuan and Zheng, Juepeng and Li, Weijia and Zhang, Wayne and others},
  journal={IEEE Transactions on Geoscience and Remote Sensing},
  year={2026},
  publisher={IEEE}
}

@inproceedings{ye2025whereami,
  title={Where am i? cross-view geo-localization with natural language descriptions},
  author={Ye, Junyan and Lin, Honglin and Ou, Leyan and Chen, Dairong and Wang, Zihao and Zhu, Qi and He, Conghui and Li, Weijia},
  booktitle={2025 IEEE/CVF International Conference on Computer Vision (ICCV)},
  pages={5890--5900},
  year={2025},
  organization={IEEE}
}

@inproceedings{ye2024cross,
  title={Cross-view image geo-localization with Panorama-BEV Co-Retrieval Network},
  author={Ye, Junyan and Lv, Zhutao and Li, Weijia and Yu, Jinhua and Yang, Haote and Zhong, Huaping and He, Conghui},
  booktitle={European Conference on Computer Vision},
  pages={74--90},
  year={2024},
  organization={Springer}
}

@article{zheng2025treecrown,
  title   = {A Review of Individual Tree Crown Detection and Delineation From
             Optical Remote Sensing Images: Current Progress and Future},
  author  = {Zheng, Juepeng and Yuan, Shuai and Li, Weijia and Fu, Haohuan and
             Yu, Le and Huang, Jianxi},
  journal = {IEEE Geoscience and Remote Sensing Magazine},
  volume  = {13},
  number  = {1},
  pages   = {209--236},
  year    = {2025},
  doi     = {10.1109/MGRS.2024.3479871},
  url     = {https://doi.org/10.1109/MGRS.2024.3479871}
}

@article{li2020integrating,
  author    = {Li, Weijia and Dong, Runmin and Fu, Haohuan and Wang, Jie and Yu, Le and Gong, Peng},
  title     = {Integrating {Google Earth} Imagery with {Landsat} Data to Improve 30-m Resolution Land Cover Mapping},
  journal   = {Remote Sensing of Environment},
  volume    = {237},
  pages     = {111563},
  year      = {2020},
  month     = feb,
  publisher = {Elsevier},
  issn      = {0034-4257},
  doi       = {10.1016/j.rse.2019.111563},
  url       = {https://doi.org/10.1016/j.rse.2019.111563}
}

@inproceedings{li2023omnicity,
  author    = {Li, Weijia and Lai, Yawen and Xu, Linning and Xiangli, Yuanbo and Yu, Jinhua and He, Conghui and Xia, Gui-Song and Lin, Dahua},
  title     = {{OmniCity}: Omnipotent City Understanding with Multi-Level and Multi-View Images},
  booktitle = {Proceedings of the IEEE/CVF Conference on Computer Vision and Pattern Recognition (CVPR)},
  pages     = {17397--17407},
  year      = {2023},
  month     = jun,
  publisher = {IEEE},
  doi       = {10.1109/CVPR52729.2023.01669},
  url       = {https://doi.org/10.1109/CVPR52729.2023.01669}
}

@article{li2023joint,
  author    = {Li, Weijia and Zhao, Wenqian and Yu, Jinhua and Zheng, Juepeng and He, Conghui and Fu, Haohuan and Lin, Dahua},
  title     = {Joint Semantic--Geometric Learning for Polygonal Building Segmentation from High-Resolution Remote Sensing Images},
  journal   = {ISPRS Journal of Photogrammetry and Remote Sensing},
  volume    = {201},
  pages     = {26--37},
  year      = {2023},
  month     = jul,
  publisher = {Elsevier},
  issn      = {0924-2716},
  doi       = {10.1016/j.isprsjprs.2023.05.010},
  url       = {https://doi.org/10.1016/j.isprsjprs.2023.05.010}
}

@inproceedings{li20243d, 
  author    = {Li, Weijia and Yang, Haote and Hu, Zhenghao and Zheng, Juepeng and Xia, Gui-Song and He, Conghui},
  title     = {{3D} Building Reconstruction from Monocular Remote Sensing Images with Multi-Level Supervisions},
  booktitle = {Proceedings of the IEEE/CVF Conference on Computer Vision and Pattern Recognition (CVPR)},
  pages     = {27728--27737},
  year      = {2024},
  month     = jun,
  publisher = {IEEE},
  doi       = {10.1109/CVPR52733.2024.02619},
  url       = {https://doi.org/10.1109/CVPR52733.2024.02619}
}

@article{dang2026skiller,
  title={SKILLER: Language-Level Reinforcement Learning for Reusable Skill Extraction in Small Language Models},
  author={Dang, Chenhao and Xiong, Siyuan and He, Conghui and Li, Weijia},
  journal={arXiv preprint arXiv:2608.10538},
  year={2026}
}

\end{document}